\documentclass[sigconf, nonacm]{acmart}
\usepackage{graphicx}
\usepackage{subcaption}

\AtBeginDocument{%
  \providecommand\BibTeX{{%
    \normalfont B\kern-0.5em{\scshape i\kern-0.25em b}\kern-0.8em\TeX}}}

\setcopyright{acmcopyright}

\copyrightyear{2020}
\acmYear{2020}
\setcopyright{acmlicensed}\acmConference[MIG '20]{Motion, Interaction and Games}{October 16--18, 2020}{Virtual Event, SC, USA}
\acmBooktitle{Motion, Interaction and Games (MIG '20), October 16--18, 2020, Virtual Event, SC, USA}
\acmPrice{15.00}
\acmDOI{10.1145/3424636.3426907}
\acmISBN{978-1-4503-8171-0/20/10}

\begin{document}

\title{
Learning to Locomote: Understanding How\\Environment Design Matters for Deep Reinforcement Learning
}

\author{Daniele Reda}
\authornote{Equal contribution.}
\email{dreda@cs.ubc.ca}
\affiliation{
  \institution{University of British Columbia}
  \city{Vancouver}
  \country{Canada}}

\author{Tianxin Tao}
\authornotemark[1]
\email{taotianx@cs.ubc.ca}
\affiliation{
  \institution{University of British Columbia}
  \city{Vancouver}
  \country{Canada}}

\author{Michiel van de Panne}
\email{van@cs.ubc.ca}
\affiliation{
  \institution{University of British Columbia}
  \city{Vancouver}
  \country{Canada}}

\begin{abstract}



Learning to locomote is one of the most common tasks in physics-based animation and deep reinforcement learning (RL).
A learned policy is the product of the problem to be solved, as embodied by the RL environment, and the RL algorithm.
While enormous attention has been devoted to RL algorithms, much less is known about the impact of design choices
for the RL environment. 
In this paper, we show that environment design matters in significant ways and document how 
it can contribute to the brittle nature of many RL results.
Specifically, we examine choices related to
 state representations,
 initial state distributions,
 reward structure,
 control frequency,
 episode termination procedures,
 curriculum usage, 
 the action space,
 and the torque limits.
We aim to stimulate discussion around such choices, which in practice strongly impact the success of RL
when applied to continuous-action control problems of interest to animation, such as learning to locomote.

\end{abstract}

\ccsdesc[500]{Computing methodologies~Animation}
\ccsdesc[500]{Computing methodologies~Reinforcement learning}

\keywords{reinforcement learning, simulation environments, locomotion}


\maketitle

\section{Introduction}
\label{introuction}

The skilled control of movement, via learned control policies, is an important problem
of shared interest across physics-based character animation, robotics, and machine learning.
In particular, the past few years have seen an explosion of shared interest in learning to locomote using 
deep reinforcement learning (RL).
From the machine learning perspective, this interest stems in part from 
the multiple locomotion tasks that found in RL benchmark suites, such as the OpenAI Gym.
This allows for a focus on the development of new RL algorithms, which
can then be benchmarked against problems that exist as predefined {\em RL environments}.
In their basic form, these environments are innocuous encapsulations of the simulated world
and the task rewards: at every time step they accept an action as input, and provide
a next-state observation and reward as output.
Therefore, this standardized encapsulation is general in nature and is well suited for benchmark-based comparisons.

However, there exist a number of issues and decisions
that arise when attempting to translate a given design intent into the canonical form of an RL environment. 
These translational issues naturally arise in the context of animation problems, where design intent is
the starting point, rather than a predefined RL environment such as found in common RL benchmarks.
In this paper, we examine and discuss the roles of the following issues related to the design of RL environments for continuous-action problems:
(a) initial-state distribution and its impact on performance and learning; 
(b) choices of state representation;
(c) control frequency or ``action repeat'';
(d) episode termination and the "infinite bootstrap" trick;
(e) curriculum learning; 
(f) choice of action space;
(g) survival bonus rewards; and
(h) torque limits.
In the absence of further understanding, these RL-environment issues contribute
to the reputation of RL as yielding brittle and unpredictable results.
Some of these issues are mentioned in an incidental fashion in work in animation and RL, which serves to motivate the more comprehensive synthesis 
and experiments presented in this paper.
We document these use cases in the sections dedicated to individual issues. 

Our work is complementary to recent RL work that examines the effect of algorithm hyperparameters, 
biases, and implementation details, 
e.g.,~\cite{zhang2018dissection, henderson2018deep, rajeswaran2017towards, packer2018assessing, ponsen2009abstraction, hessel2019inductive, andrychowicz2020matters}.
\citet{henderson2018deep} discusses the problem of reproducibility in RL, due to extrinsic factors, such as hyperparameter selections and different code bases, and intrinsic factors, such as random seeds and implicit characteristics of environments.
\citet{hessel2019inductive} analyze the importance of multiple inductive biases and their impact on the policy performance, including discount, reward scaling and action repetitions. 
\citet{Engstrom2020Implementation} points to certain code-level optimizations that sometimes contribute the bulk of performance improvements. Some of these configuration decisions are more impactful than others. The performance of RL policies can be sensitive to the change of hyperparameters which indicates the brittleness of RL algorithms.
\citet{andrychowicz2020matters}, concurrent to our work, performs a large scale ablation of hyperparameters and implementation choices for on-policy RL, including a mix of algorithmic and environment choices.

\section{Experimental Setting}
\label{experiments}

We consider the traditional RL setting, in which a Markov Decision Process (MDP) is defined by
a set of states $\mathcal{S}$,
a set of actions $\mathcal{A}$,
a transition probability function $p \colon \mathcal{S} \times \mathcal{A} \to \mathcal{P}(\mathcal{S})$, which assigns to every pair $(s, a) \in \mathcal{S} \times \mathcal{A}$ a probability distribution $p(\cdot | s, a)$, representing the probability of entering a state from state $s$ using action $a$,
a reward function $R \colon \mathcal{S} \times \mathcal{S} \times \mathcal{A} \to \mathbb{R}$, which describes the reward $R(s_{t+1}, s_t, a_t)$, associated with entering state $s_{t+1}$ from state $s_t$, using action $a_t$,
and a future discount factor $\gamma \in [0, 1]$ representing the importance of future rewards.
The solution of an MDP is a policy $\pi_\phi \colon \mathcal{S} \to \mathcal{A}$, parameterized by parameters $\phi$, that for every $s_0 \in \mathcal{S}$ maximises:
$J(\phi) = \mathbb{E} \left[ \sum_{t=0}^{\infty} \gamma^{t} R(s_{t+1}, s_t, a_t) \right]$,
where the expectation is taken over states $s_{t+1}$ sampled according to $p(s_{t+1} | s_t, a_t)$ and $a_t$ is the action sampled from $\pi_\phi (s_t)$.

Our algorithm of choice for the following experiments is TD3~\cite{fujimoto2018addressing}, a state of the art, off-policy, model free, actor-critic, RL algorithm. In TD3, an actor represents the policy $\pi_\phi$ where $\phi$ are the weights of a neural network optimized by taking the gradient of the expected return $\nabla_{\phi} J(\phi)$, through the deterministic policy gradient algorithm~\citep{silver2014deterministic}:
$\nabla_{\phi} J(\phi) = \mathbb{E} \left[ \nabla_a Q^\pi(s,a) \nabla_{\phi} \pi_\phi(s) \right]$, and a critic is used to approximate the expected return $Q^\pi(s,a)$ which corresponds to taking an action $a$ in state $s$ and following policy $\pi$ thereafter. 
The critic is also a neural network parameterized by $\theta$ and optimized using temporal difference learning using target networks~\cite{mnih2015human} to maintain a fixed objective $\nabla_\theta (Q^\pi_\theta(s, a) - y)^2$ where $y$ is the target and is defined as $y = r_t + \gamma \left[ Q_{\theta target} (s_{t+1}, \pi_\phi(s_{t+1})) \right]$. 

We explore, evaluate, and discuss the impact of each defining component of the environment, each potentially affecting the final performance and learning efficiency of policies in locomotion environments. For each of these experiments, unless otherwise stated, we use an implementation of TD3 based on the original code from the authors and adapted for more experimental flexibility.
Hyperparameters and network architectures are given in appendix~\ref{appendix:hyperparams}. All of our experiments are based on the Bullet physics simulator~\cite{coumans2019} and its default Roboschool locomotion environments available through Gym~\cite{openaigym}. Each experiment is performed a total of 10 times, and averaged across seeds. The set of seeds is always the same.
In a few cases, we rely on existing published results, using them to provide relevant insights in our context.

We now present each RL environment design issue in turn. For clarity, we combine the explanation, related work, and results for each.

\section{Initial-state Distribution}
\label{initial_state}

\begin{figure*}
\centering
\includegraphics[width=\textwidth]{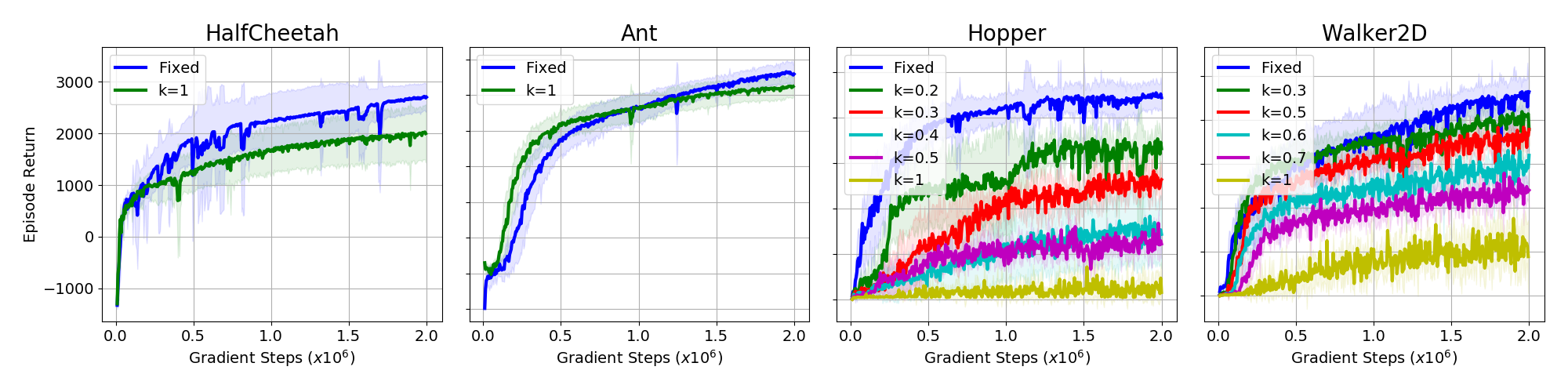}

\caption{Learning curves as a function of the initial-state distribution parameterized by $\kappa$. For the Hopper and Walker environments, multiple values between 0 and 1 are sampled to better illustrate the relationship between the initial-state distribution and training performance.}
\label{fig:random_initial_results}
\end{figure*}

\begin{table}[tbh]
\begin{center}
\begin{small}
\begin{tabular}{lc|cc}
\toprule
\bf{Test environment}        & \bf{$\kappa$} & \bf{Matching ISD} &\bf{Narrow ISD}\\
\midrule
AntBulletEnv            & 0               & 3422.26                    & N/A\\
AntBulletEnv            & 1               & 3139.38                    & 1647.18\\
\hline
HalfCheetahBulletEnv    & 0               & 2808.28                    & N/A\\
HalfCheetahBulletEnv    & 1               & 2001.52                    & 2117.88\\
\hline
HopperBulletEnv         & 0               & 2300.24                    & N/A\\
HopperBulletEnv         & 0.2             & 1622.55                    & 1286.58\\
HopperBulletEnv         & 0.3             & 1357.53                    & 761.47\\
HopperBulletEnv         & 0.4             & 995.27                     & 396.90\\
HopperBulletEnv         & 0.5             & 528.76                     & 325.70\\
HopperBulletEnv         & 1               & 528.76                     & 325.70\\
\hline
Walker2DBulletEnv       & 0               & 2459.30                    & N/A\\
Walker2DBulletEnv       & 0.3             & 2358.83                    & 644.02\\
Walker2DBulletEnv       & 0.5             & 1792.50                    & 199.11\\
Walker2DBulletEnv       & 0.6             & 1497.13                    & 131.47\\
Walker2DBulletEnv       & 0.7             & 1255.69                    & 60.60\\
Walker2DBulletEnv       & 1               & 584.91                     & 9.78\\
\bottomrule
\end{tabular}
\end{small}
\end{center}
\caption{\label{tab:policy_eval}Initial state distributions (ISD) generalization experiments. Here, Matching ISD gives the test reward when the training and test environments are matched in terms of their initial state distribution. Narrow ISD gives the test rewards when trained on a narrow distribution ($\kappa$=0), and tested on the various broader initial state distributions.}
\end{table}

The distribution from which the initial states of simulation episodes are sampled from plays an important role in learning and in exploration in RL. In many reinforcement learning locomotion benchmarks~\cite{todorov2012mujoco, coumans2019}, the default initial state is sampled from a very narrow distribution, nearly deterministic, at the beginning of a new episode. 
Recent work~\cite{ghosh2018divideandconquer} addresses the problem of solving harder tasks by partitioning the initial state space into slices and train an ensemble of policies, one on each slice.
\citet{packer2018assessing} looks at a similar problem, but instead of modifying the initial-state distribution, changes the environment dynamics, by changing mass, forces and lengths of links at every trajectory. In our experiments, changing the initial-state distribution does not affect the underlying environment dynamics, which remain the same. It instead affects the way the state-space is presented to the RL algorithm.

Default PyBullet locomotion environments create a new initial state by sampling the joint angles of each link from a uniform distribution $\mathcal{U}(-0.1, 0.1)$. Instead, we sample each joint angle from a new distribution $\mathcal{U}(\kappa \theta_{\min}, \kappa \theta_{\max})$, where $\theta_{\min}$ and $\theta_{\max}$ are the specified lower and upper limit of the joint, predefined in the robot description files (i.e. usually URDF or XML files). 
The parameter $\kappa$ quantifies the width of the initial-state distribution to investigate how its variance affects the policy. 

We run experiments with the following environments: Ant, HalfCheetah, Hopper and Walker2D, and illustrate the test episode return in Figure~\ref{fig:random_initial_results}.
For Ant and HalfCheetah environments, results for the original narrow initial-state distribution and the broad initial-state distribution with $\kappa=1$ are shown.
For the Hopper and Walker environment, we show additional training curves with different values of $\kappa$ to investigate the influence of $\kappa$ at finer scales. 
A broader initial-state distribution leads to worse sampling efficiency and a lower episode return in the end for the majority of environments.
We believe that for Ant and HalfCheetah, the largely invariant nature of the learning 
to the initial state distributions stems from the large degree of natural static stability 
for these systems after falling to the ground. 
This stability leads to a rapid convergence to similar states from a wide range of initial states. 
 
The difference in performance caused by broadening the initial-state distribution reflects its impact, which is often neglected, in policy training. A narrow initial-state distribution increases learning efficiency compared with tasks with same underlying mechanics $P(s_{t+1}|s_{t},a_{t})$ but a broader initial-state distribution $p(s_{0})$. Intuitively, having a restricted range of initial states allows the agent to focus on those initial states, from which it learns how to act. If the agent is always dropped in the environment in a very different state, it is more difficult for the policy to initially learn to act, given that the experiences 
will represent very disjoint regions of the state space. 
However, learning on a broader initial-state distribution results in a more robust and general final policy.
Table~\ref{tab:policy_eval} shows results for policies trained on the default environments, with narrow initial-state distribution, when tested with different initial-state distributions. The policies trained on narrow initial-state distributions deliver significant worse run-time performances as $\kappa$ increases, showing  a failure to generalize with respect to much of the state space. Thus, training with a broader initial-state distribution lead to a more robust policy that covers a wider range of the state and action space.

\textbf{Summary:} RL results can be strongly impacted by the choice of initial state distribution. 
A policy trained with the narrow initial distribution in commonly used environments 
often fails to generalize to regions of the state space that likely remain unseen during training. 
The broader the initial state distribution $p_0(s)$, the more difficult the RL problem can become. 
As a result, there is a drop in both sample-efficiency and reward with a wider initial-state distribution. 
However, the learned policies are more generalizable and more robust.

\section{State Representation}
\label{starting_state}

The state, at any given time, captures the information needed by the policy in order to understand the current condition of the environment and decide what action to take next. Despite the success of end-to-end learning, the choice or availability in the state representation can still affect the problem difficulty.
In this section, we investigate how adding, removing and modifying information in the state affects the RL benchmarks.
We discuss the following modifications:
 (1) adding a phase variable for cyclic motions,
 (2) augmenting the state with the Cartesian joint position in Cartesian coordinate,
 (3) removing contact boolean variables,
 (4) using the initial layers of a pre-trained policy as state representation for a new policy.

\subsection{Phase Variable}

\begin{figure}
\begin{subfigure}{0.48\columnwidth}
  \centering
  \includegraphics[width=\columnwidth]{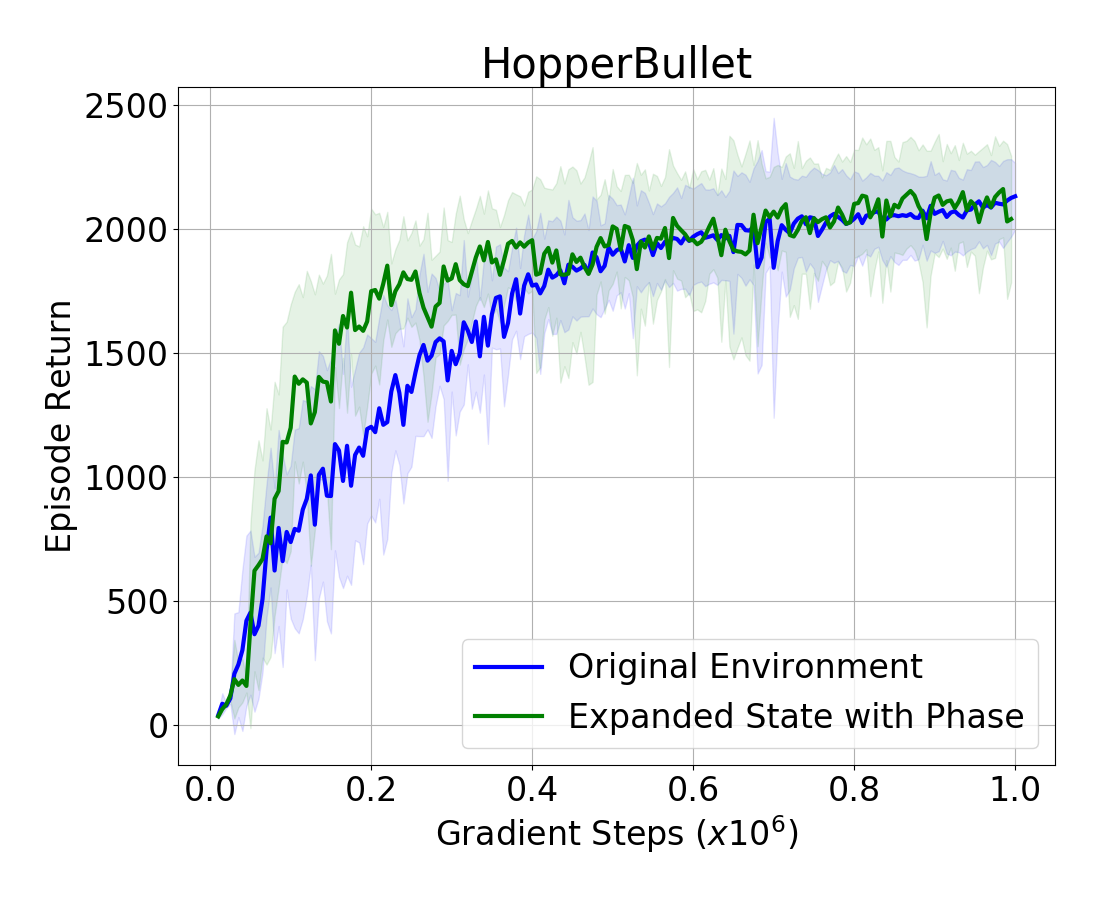}
  \caption{HopperBulletEnv}
  \label{fig:hopper_phase}
\end{subfigure}
\begin{subfigure}{0.48\columnwidth}
  \centering
  \includegraphics[width=\columnwidth]{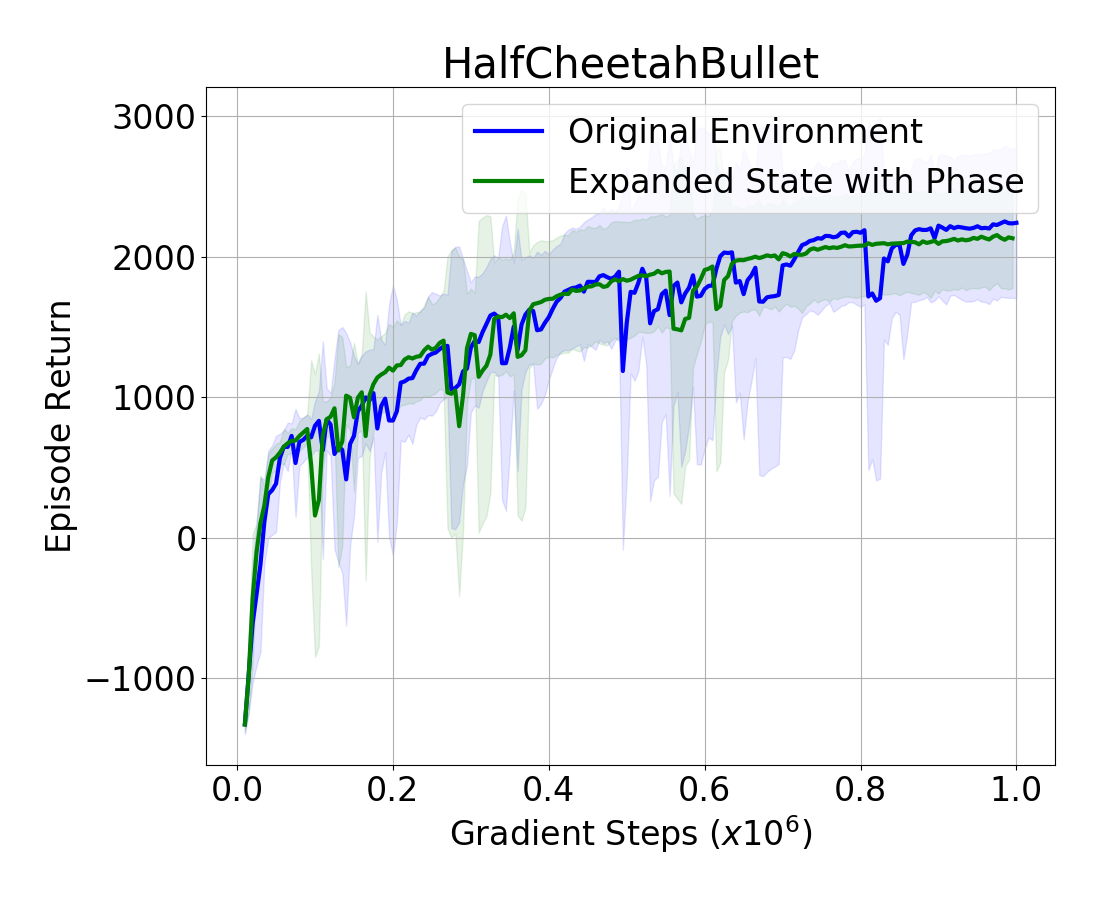}  
  \caption{HalfCheetahBulletEnv}
  \label{fig:halfcheetah_phase}
\end{subfigure}
\caption{Performance of TD3 with sin and cos of the phase variable added to the state description. Blue: training curve of the original environment. Green: training curve of the environment with state variable.}
\label{fig:phase_results} 
\end{figure}

Previous work on motion-imitation based control, e.g., ~\cite{2018-TOG-deepMimic, 2019-MIG-symmetry} 
uses a phase variable $\phi \in [0, 2\pi]$ as part of the state representation to implicitly index 
the desired reference pose to imitate at a given time.
We can use prior knowledge about the desired period of the motion, $T$, to define a phase variable,
even when there is no specific motion to imitate:
\begin{equation*}
    \phi = (\frac{2\pi}{T})t,~ 0 \leq t < T
\end{equation*}
The phase variable, $\phi$, can assist mastering periodic motion because it provides a compact abstraction of the state variables for nearly cyclic motions. Ideally, given a precise phase value for periodic motion for a specific character, the RL algorithm should perform better since the phase variable acts as a state machine to guide the learning process.
We estimate $\phi$ from a previously learned controller and include this in the state representation.

Phase has been shown to serve as an effective internal representation for kinematic 
motion synthesis in Phase-Functioned Neural Networks (PFNN)~\cite{holden2017phase}.
They study the impact of phase by comparing three cases: using phase as an input to a gating network; 
using phase as an input to a fully connected network; and a baseline case, e.g., fully connected network without phase. 
The gating network with phase input provides the best motion quality, with crisp motion and ground contacts.
Without a phase input, the simulated characters exhibit strong foot-skate artifacts,
while a fully connected network with phase input often ignores the phase input, similarly resulting in lower-quality motions.

To prevent the phase variable being ignored, and inspired by the solution proposed by~\cite{hawke2019urban} for the control command, we input the phase variable at every layer. We expect this modification to encourage the network to use these additional input features while both actor and critic networks still remain functional.
Instead of feeding the phase variable $\phi$ to the actor and critic networks, we input $sin(\phi)$ and $cos(\phi)$. 

The averaged episode return over 10 runs are plotted in Figure~\ref{fig:phase_results}. In the Hopper environment, the episode return for the state with phase has a more efficient learning curve, but eventually it converges to a similar result at the end. This indicates the fact that adding phase information to training can accelerate the learning process and result in better sampling efficiency. However, phase information does not lead to a higher return and this is explained by the fact that phase only provides an abstraction of the joint angles and velocities, adding no extra information that is not already present in the rest of the state. This can be further seen in the HalfCheetah environment, where there is no improvement when adding the phase information. 

\textbf{Summary: }Adding phase information has limited impact on learning locomotion in the benchmark environments, 
which do not use reference motions but that could in principle benefit from a basic form of periodicity, such as that provided by a phase variable.

\subsection{Joint Position}

\begin{figure}
\begin{subfigure}{0.48\columnwidth}
  \centering
  \includegraphics[width=\columnwidth]{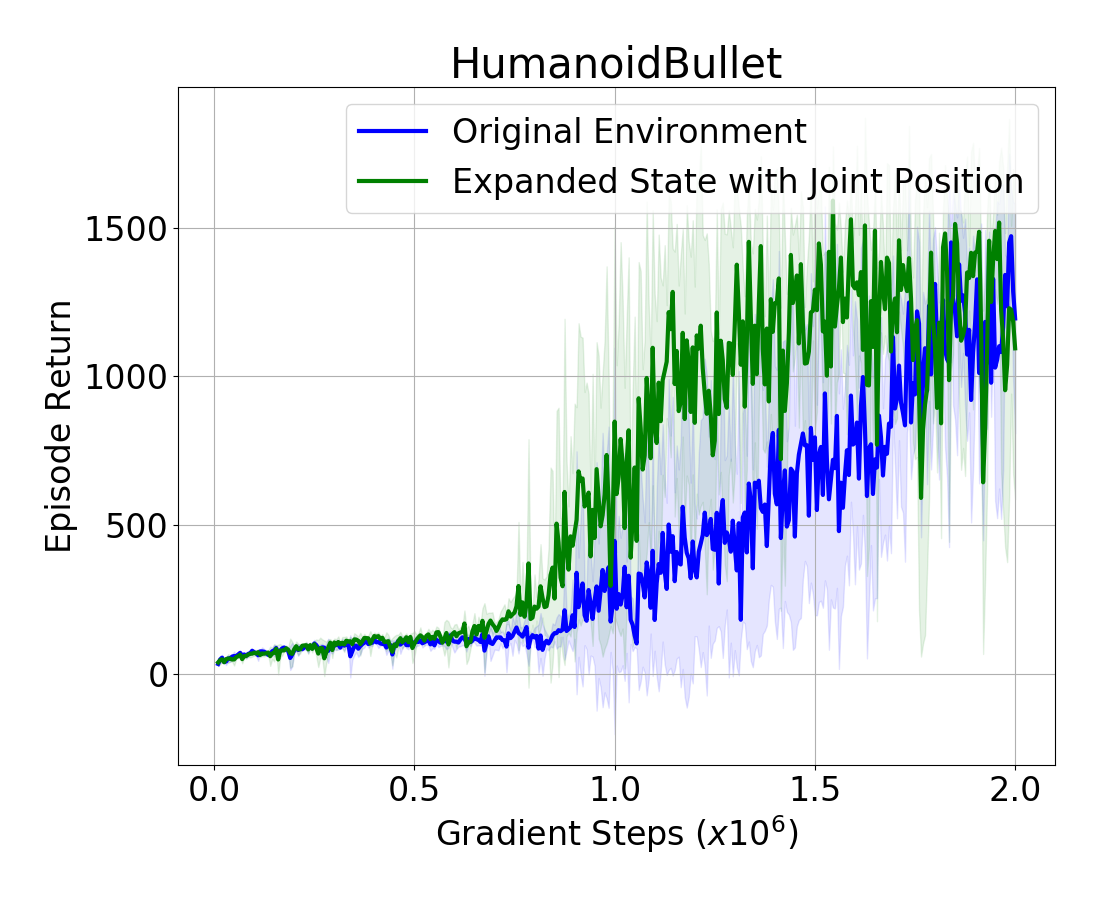}
  \caption{HumanoidBulletEnv}
  \label{fig:humanoid_jointpos}
\end{subfigure}
\begin{subfigure}{0.48\columnwidth}
  \centering
  \includegraphics[width=\columnwidth]{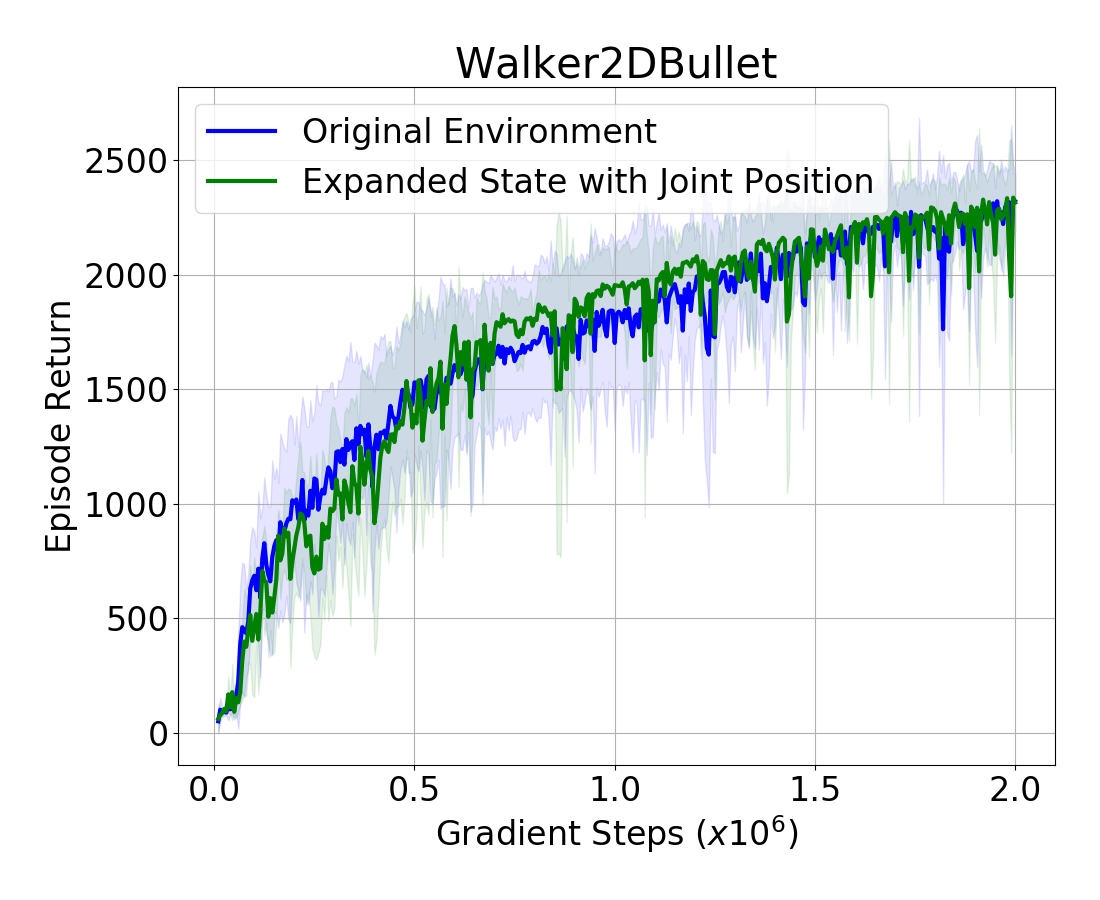}  
  \caption{Walker2DBulletEnv}
  \label{fig:walker_jointpos}
\end{subfigure}
\caption{Training curves for a state representation augmented with joint positions.} 
\label{fig:joint_pos_results} 
\end{figure}

The inclusion of Cartesian joint position in the state description is another major difference between the work from the computer graphics community and RL community. In previous studies on motion synthesis and motion imitation~\cite{zhang2018mode,2018-TOG-deepMimic,holden2017phase,Park:2019}, both joint angles and Cartesian joint position are included in the state description. Although the joint positions provide redundant information, they are still found to be helpful in learning kinematic motion models and physics-based controllers. 
We expect it to be more important for complex 3D characters because of the utility of knowing where the feet are with respect to the ground and the potential complexity of the forward kinematics computation needed to compute this.
Here we investigate the use of Cartesian joint position for learning physics-driven locomotion. The Cartesian joint positions are represented in a Cartesian coordinate system centered at the root of the character.


We train with state representations that are augmented with Cartesian joint position, for the 3D Humanoid  and Walker2D environments,
and compare them against the results trained with the original state representations.
As shown in Figure~\ref{fig:joint_pos_results}, the Humanoid learns faster with the 
augmented state representation, and yields the same final performance.
The augmented state representation has negligible benefits for the Walker2D environment.

\textbf{Summary: }For more complex characters, adding Cartesian joint position to the state 
improves learning speed.

\subsection{Contact Information}

\begin{figure}
\begin{subfigure}{0.48\columnwidth}
  \centering
  \includegraphics[width=\columnwidth]{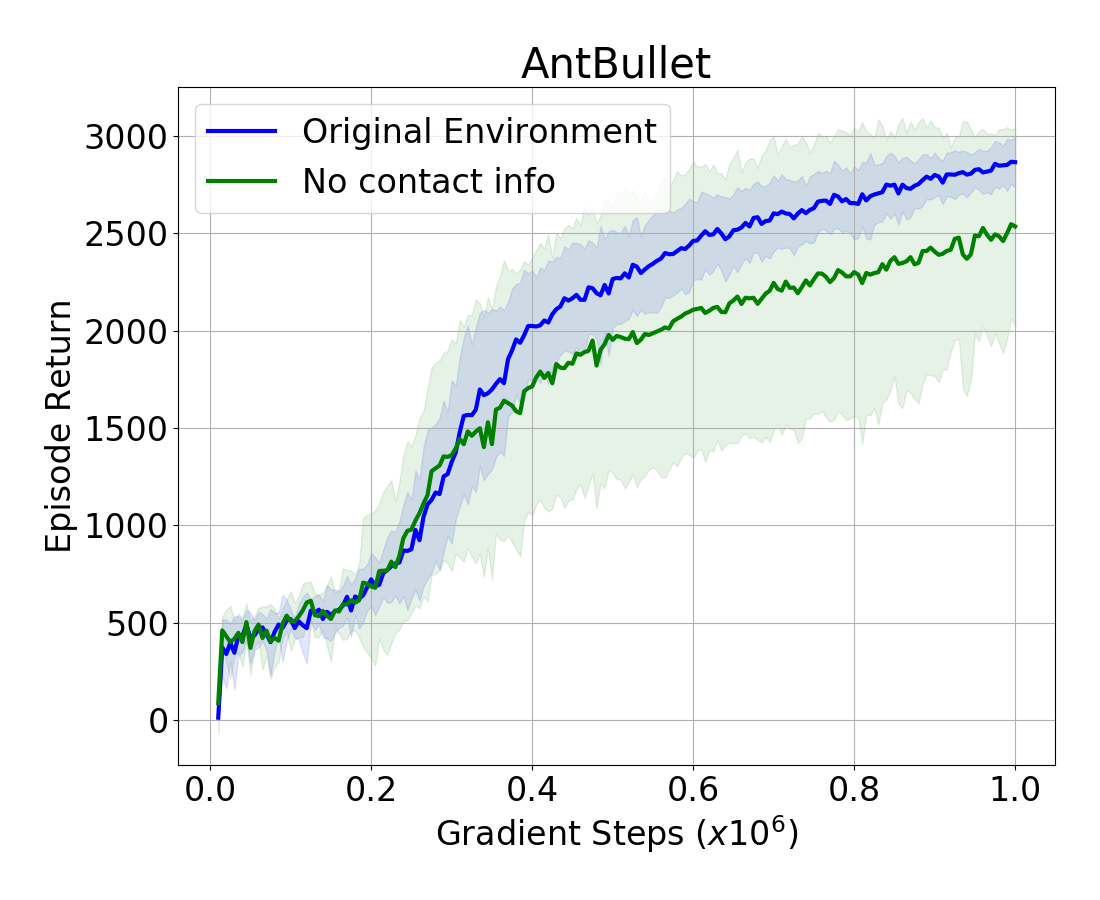}
  \caption{AntBulletEnv}
  \label{fig:ant_no_contact}
\end{subfigure}
\begin{subfigure}{0.48\columnwidth}
  \centering
  \includegraphics[width=\columnwidth]{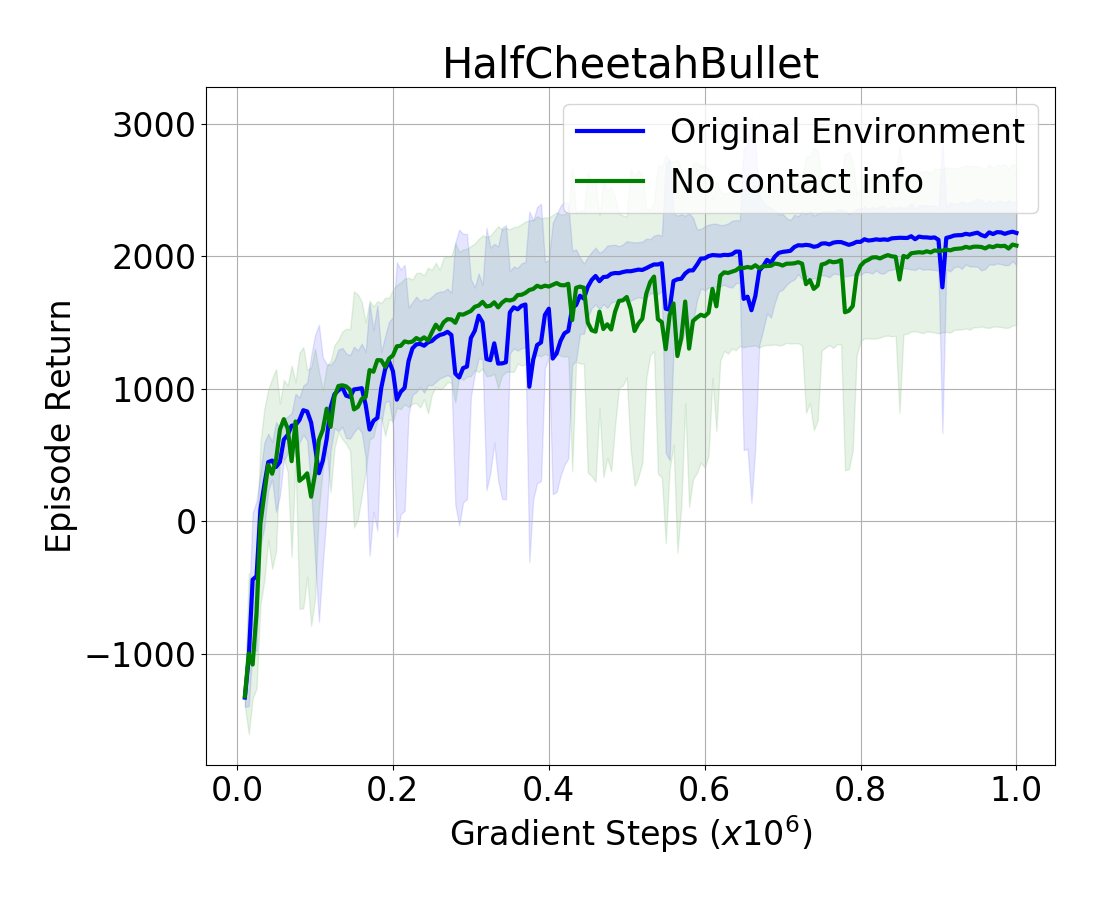}  
  \caption{HalfCheetahBulletEnv}
  \label{fig:halfcheetah_no_contact}
\end{subfigure}
~

\begin{subfigure}{0.48\columnwidth}
  \centering
  \includegraphics[width=\columnwidth]{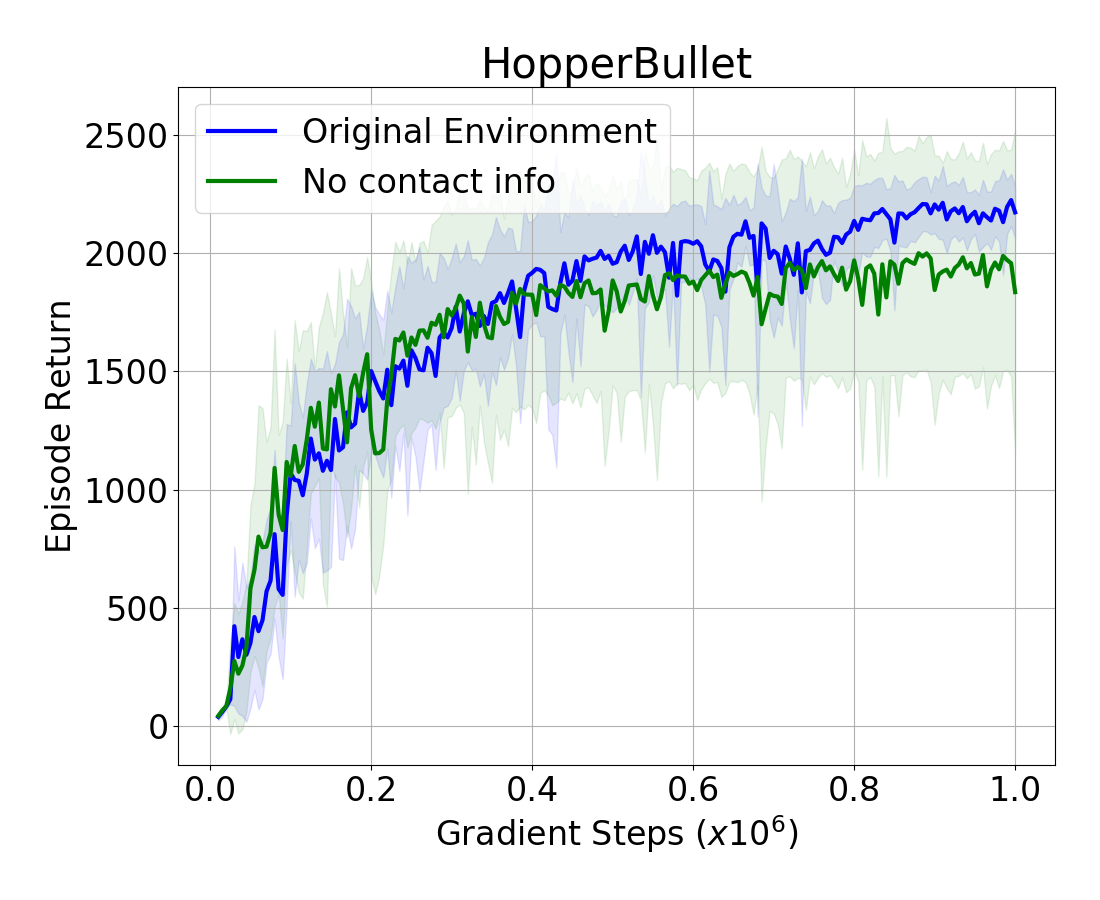}  
  \caption{HopperBulletEnv}
  \label{fig:hopper_no_contact}
\end{subfigure}
\begin{subfigure}{0.48\columnwidth}
  \centering
  \includegraphics[width=\columnwidth]{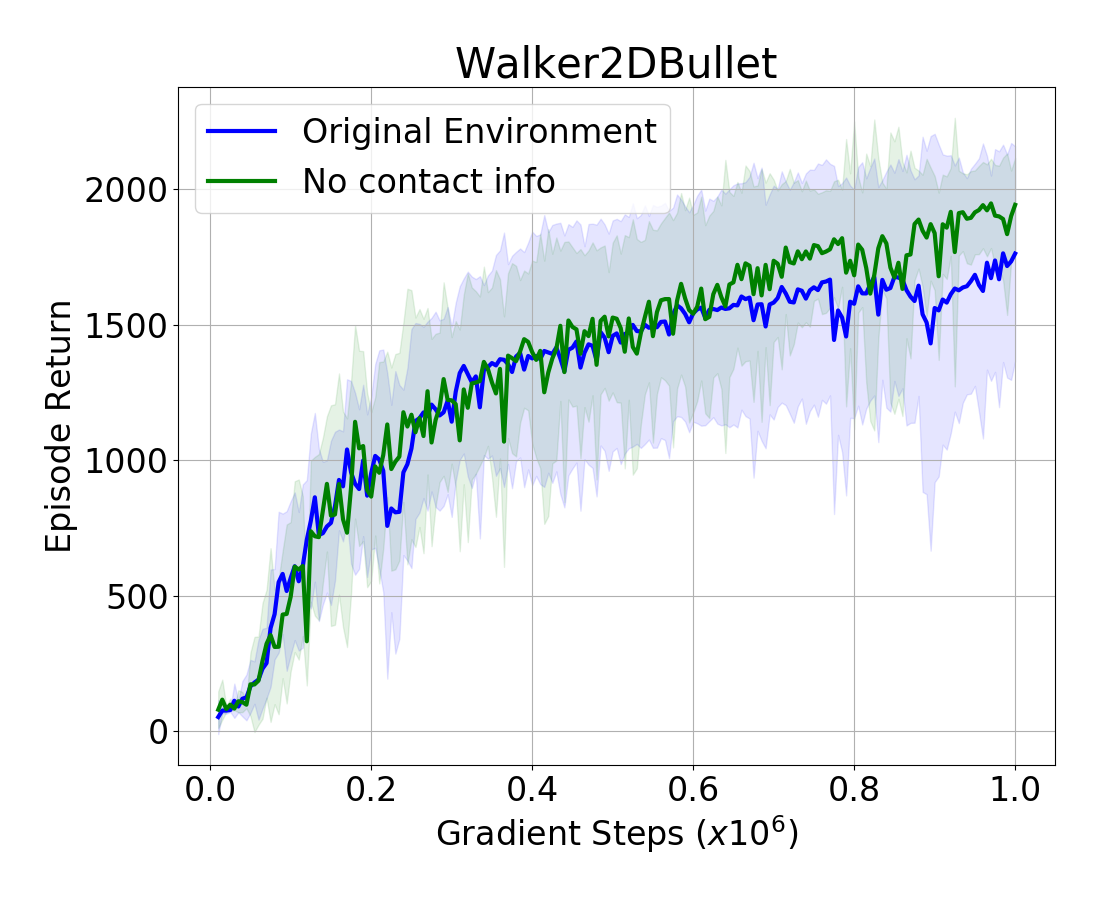}  
  \caption{Walker2DBulletEnv}
  \label{fig:walker_no_contact}
\end{subfigure}
\caption{Training curves for a state representation augmented with link contact information.}
\label{fig:no_contact_results} 
\end{figure}

The PyBullet default state includes the binary valued contact state information about link contacts with the ground.
The other popular physics simulator for locomotion environment, Mujoco~\cite{todorov2012mujoco}, does not include contact link variables.
We ask the question: \textit{Is contact information really useful?} The experimental results for HalfCheetah, Walker2D, Hopper and Ant environments are shown in Figure~\ref{fig:no_contact_results}.
For Ant, removing contact information from the state affects the final reward, making it worse, as this environment has a relatively high number of points of contact with the ground (six). For Hopper, Walker2D and HalfCheetah, one, two and four points of contact respectively, the final return is not affected.
Other ways of representing contact information, e.g., contact forces, could produce different outcomes.

\textbf{Summary:} Adding binary contact information to state variable can be helpful in some cases, 
but may have negligible benefits in other cases.

\subsection{Pre-trained Representations}

\begin{figure}
\begin{subfigure}{0.48\columnwidth}
  \centering
  \includegraphics[width=\columnwidth]{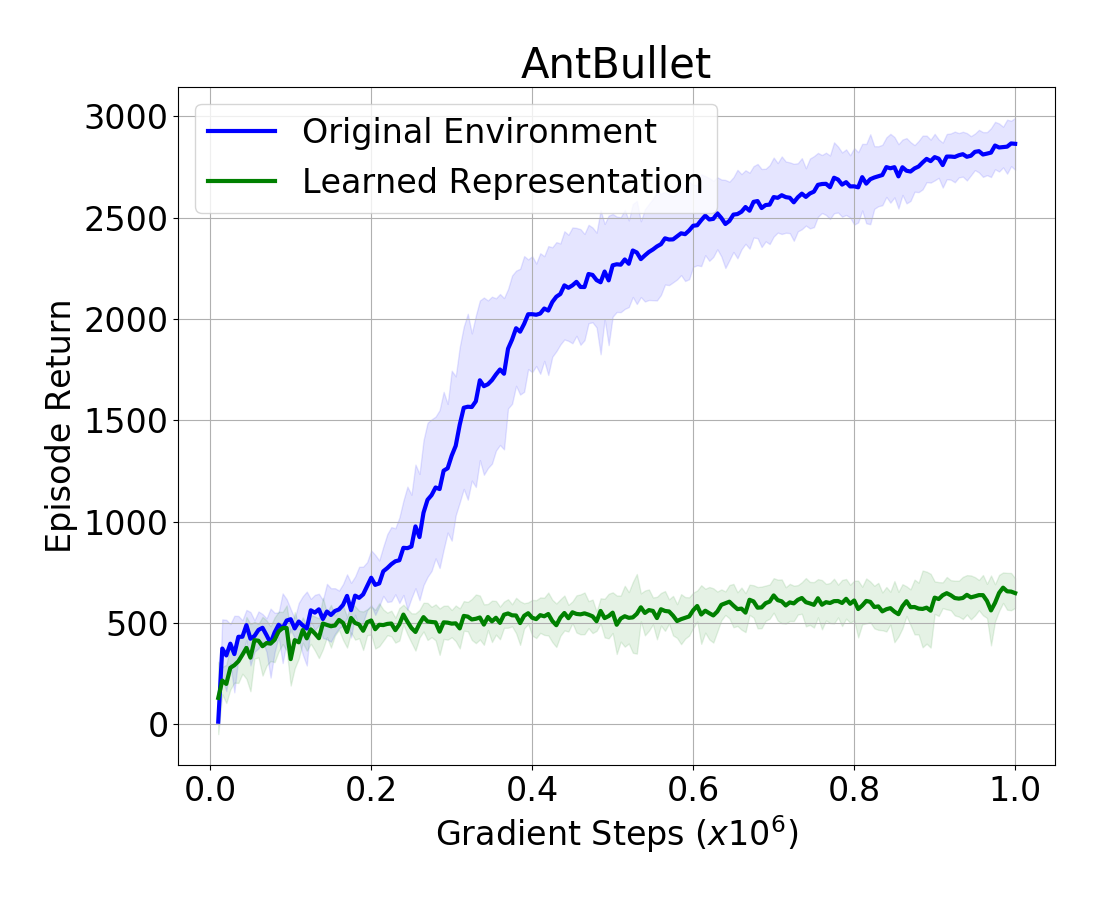}
  \caption{AntBulletEnv}
  \label{fig:ant_learned_representation}
\end{subfigure}
\begin{subfigure}{0.48\columnwidth}
  \centering
  \includegraphics[width=\columnwidth]{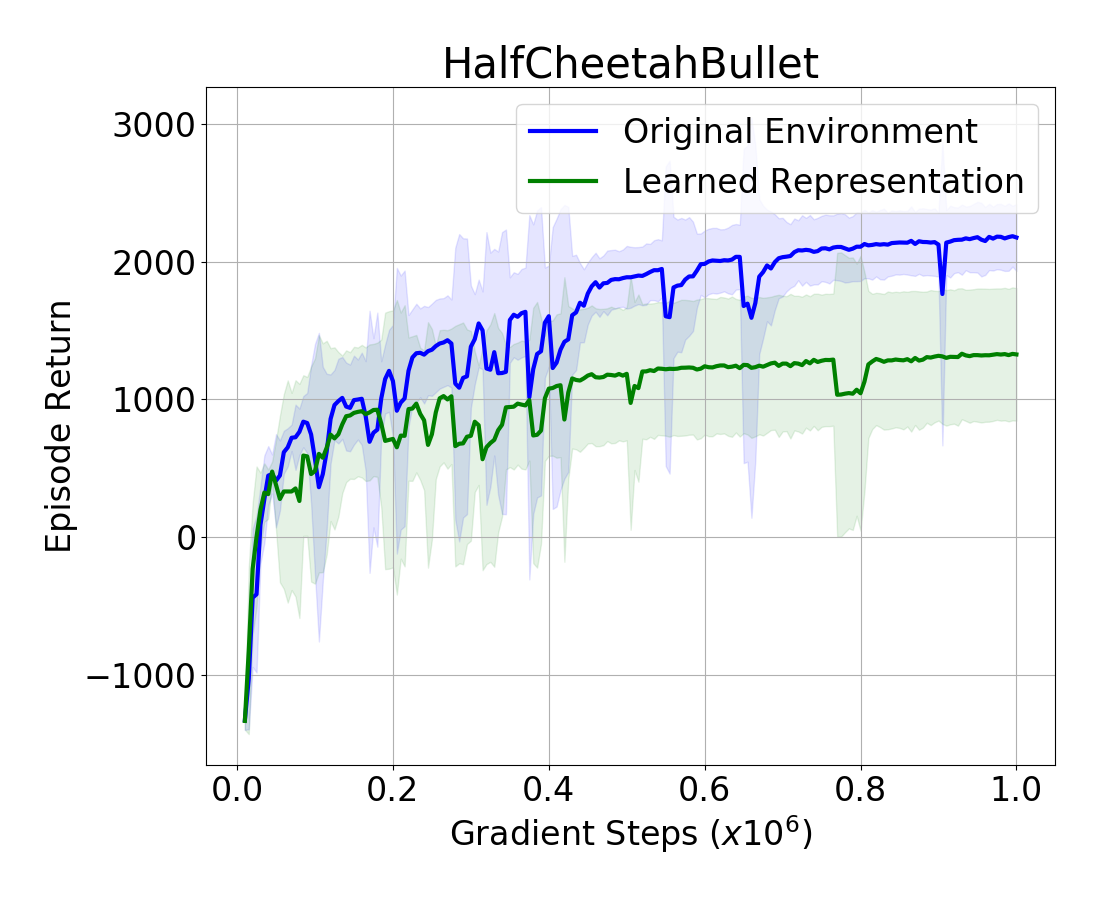}  
  \caption{HalfCheetahBulletEnv}
  \label{fig:halfcheetah_learned_representation}
\end{subfigure}
~

\begin{subfigure}{0.48\columnwidth}
  \centering
  \includegraphics[width=\columnwidth]{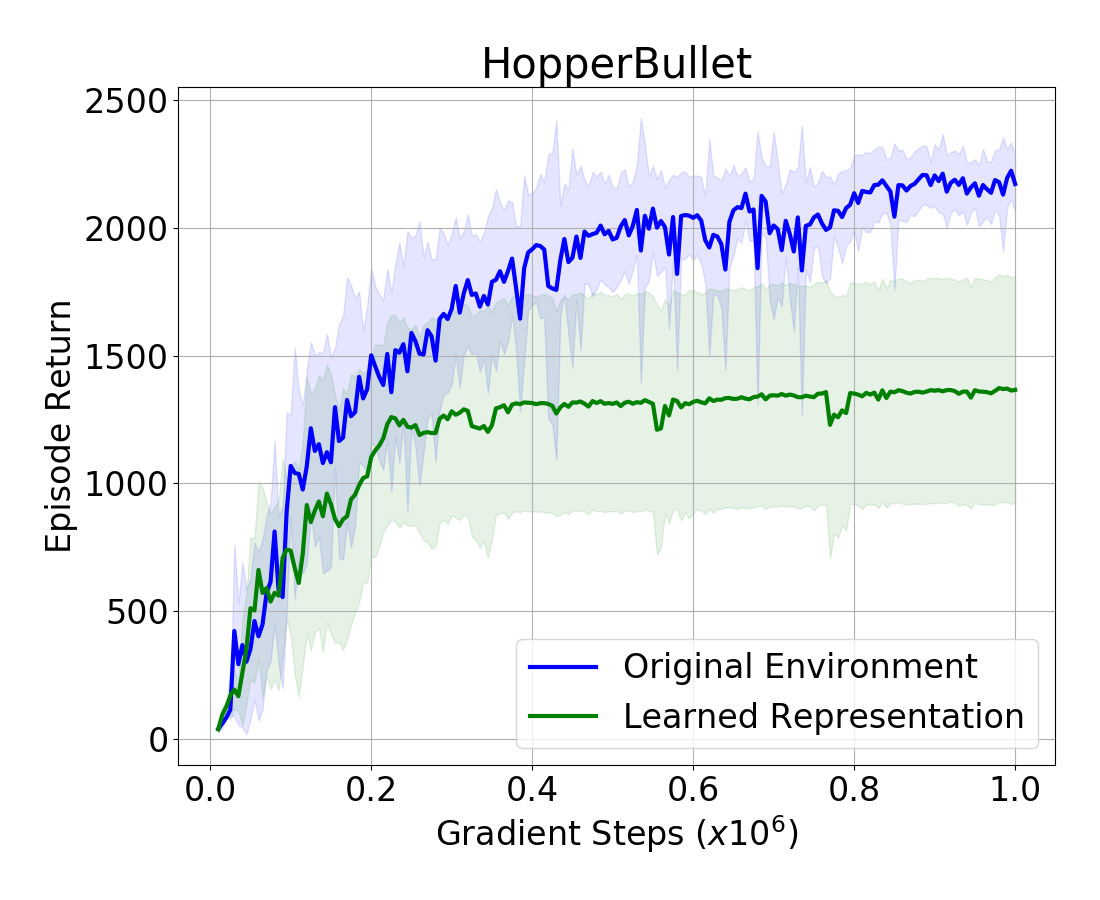}  
  \caption{HopperBulletEnv}
  \label{fig:hopper_learned_representation}
\end{subfigure}
\begin{subfigure}{0.48\columnwidth}
  \centering
  \includegraphics[width=\columnwidth]{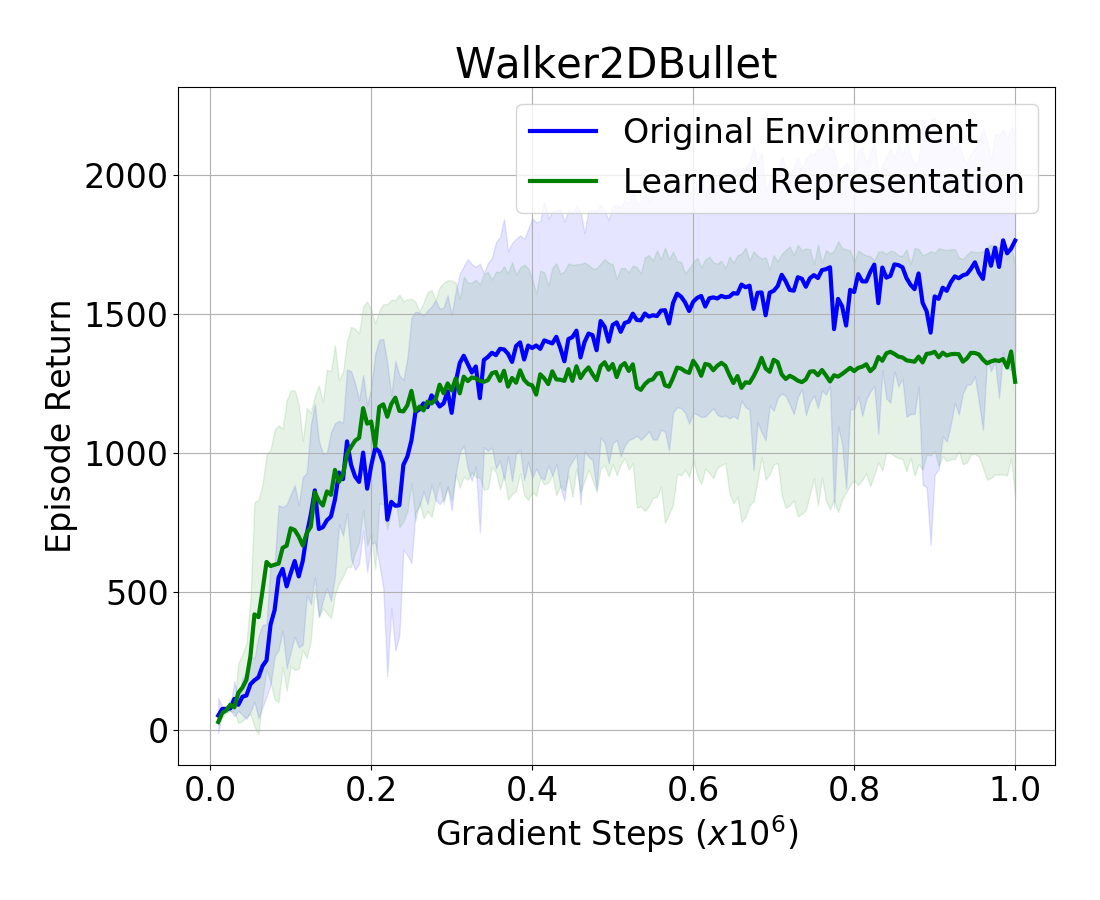}  
  \caption{Walker2DBulletEnv}
  \label{fig:walker_learned_representation}
\end{subfigure}
\caption{Learning with a pre-trained state representation.}
\label{fig:learned_repr_results} 
\end{figure}

When the observation given by the environment is high-dimensional, pixel-based for example, it is typical to learn a more compact representation that is then used for control, whether end-to-end, through pre-training or in a 2 step precedure.
When a compact state is provided, in principle a neural network transforms multiple time the state to build, layer after layer, a increasingly more suitable representation for control. In this scenario, the representation immediately before the output layer is the most suited.
Inspired by the usage of pre-trained networks, we applied this to the original compact state of Ant, HalfCheetah, Hopper and Walker2D environments.

After training a full end-to-end policy with TD3, we take the policy network and remove the output layer. We obtain a network that transforms the compact state representation into a representation of the size of the last hidden layer, and we use this as input representation of our new experiment.
Interestingly, shown in Figure~\ref{fig:learned_repr_results}, we discover that this setup provides a sub-optimal representation, which learns as efficiently as the original state initially, but plateaus earlier.
This is surprising, given that the similar approach of using a pre-trained representation is widely used with pixel-based states, but explainable by the fact that the final trained policy used as representation has steered towards the relevant part of the state space and so is unable to represent properly more exploratory states.

\textbf{Summary:} Unlike the success of pre-trained network in computer vision tasks, the pre-trained policy without the last layer does not provide a state representation easier to learn from than the raw state variable.
\section{Control Frequency}

The choice of \textit{action repeat (AR)}, also called \textit{frame skip}, has been fundamental to early work in DQN~\cite{mnih2015human} and it has been overlooked as a simple hyperparameter in many other algorithms. 
This relevant domain knowledge consists of repeating each action selected by the agent a fixed number of times. 
This effectively lowers the frequency at which the agent operates, compared to the environment frequency. 
The impact of the action repeat parameter has been studied by~\cite{hessel2019inductive}, 
in which they train an adaptive agent that tunes the action repetition. 
In~\cite{lakshminarayanan2017dynamic}, the authors propose a simple-but-limited approach that improves DQN. 
The policy outputs both an action and an action-repeat value, which is selected from one of two
possible values. Tuning these (hyperparameter) values is highly beneficial to the learning. 
More recently, \citet{metelli2020control} also explore the choice of control frequency and \textit{action persistency} for low-D continuous environments, such as CartPole and MountainCar.

We systematically examine the effect of action repeat in common PyBullet locomotion environments.
The environment frequency is fixed and given by the simulator. With $AR=1$, the control frequency and the environment frequency are equal, and a gradient step is computed after each control step. By choosing a different action repeat $AR$ value, we are scaling down the control frequency and taking a gradient step after $AR$ environment steps.
The learning curves are shown in~Figure~\ref{fig:action_repeat_results}.
As expected, the choice of action repeat
has a significant impact on the learning. For the Walker and Hopper, AR=1 produces the best performance, with larger
values producing worse performance. This also leaves open the possibility that a higher frequency control rate 
could improve learning. Humanoid learns best with AR=3 or AR=4, and  Ant with AR=2.
An exception is HalfCheetah, which is not particularly sensitive to the choice of action repeat.

\textbf{Summary:} A good choice of control frequency, usually implemented as action repeat parameter for RL, 
is essential for good learning performance. Controlling the actions at frequencies that are too high or too low
is usually harmful.

\newcommand{\figscale}{0.48\columnwidth}
\begin{figure}[ht]
\begin{subfigure}{\figscale}
  \centering
  \includegraphics[width=\columnwidth]{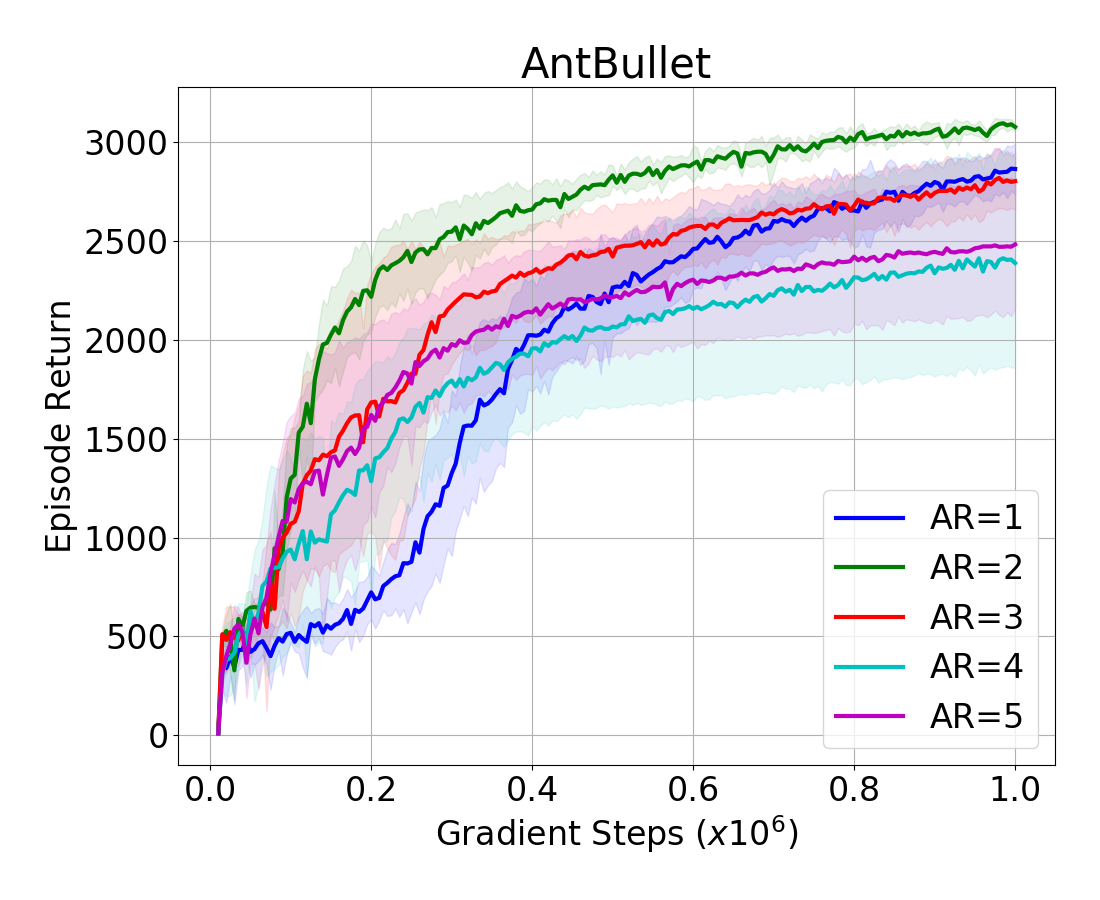}
  \caption{AntBulletEnv}
  \label{fig:ant_action_repeat}
\end{subfigure}
\begin{subfigure}{\figscale}
  \centering
  \includegraphics[width=\columnwidth]{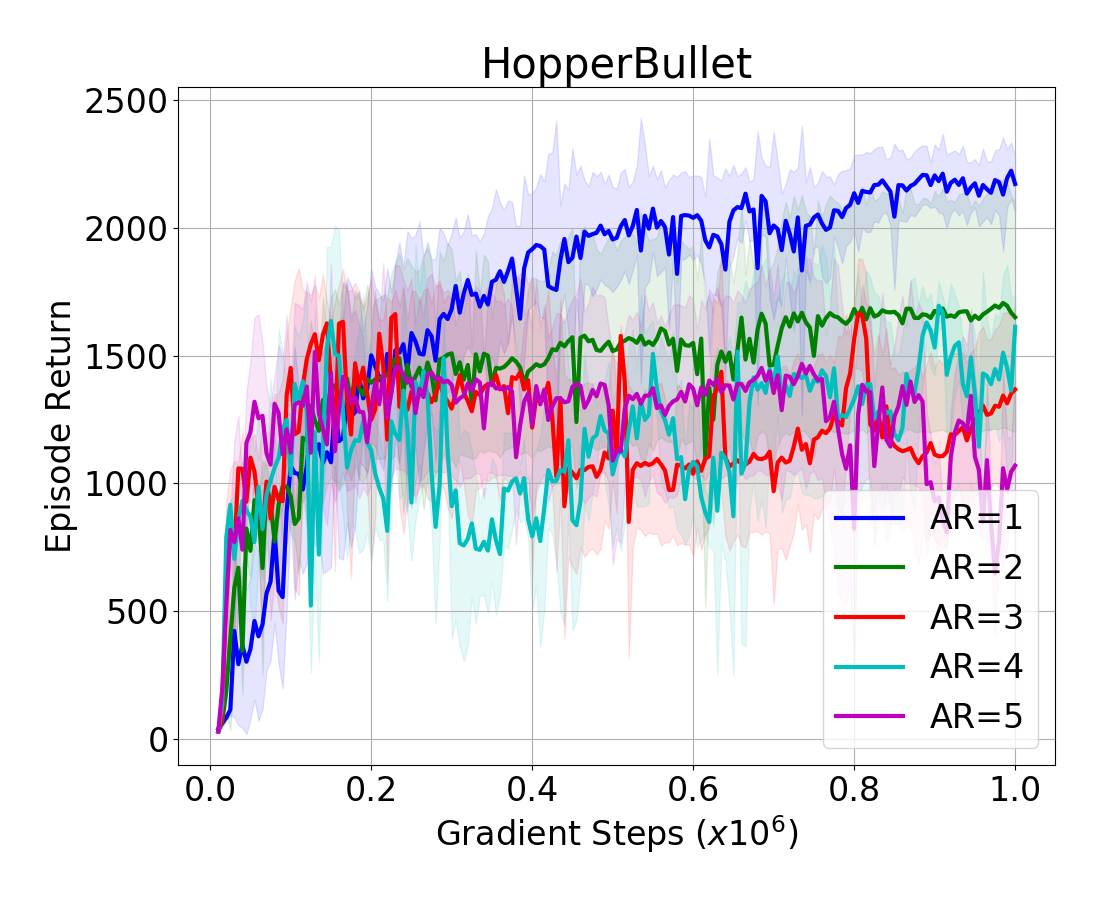}  
  \caption{HopperBulletEnv}
  \label{fig:hopper_action_repeat}
\end{subfigure}
\begin{subfigure}{\figscale}
  \centering
  \includegraphics[width=\columnwidth]{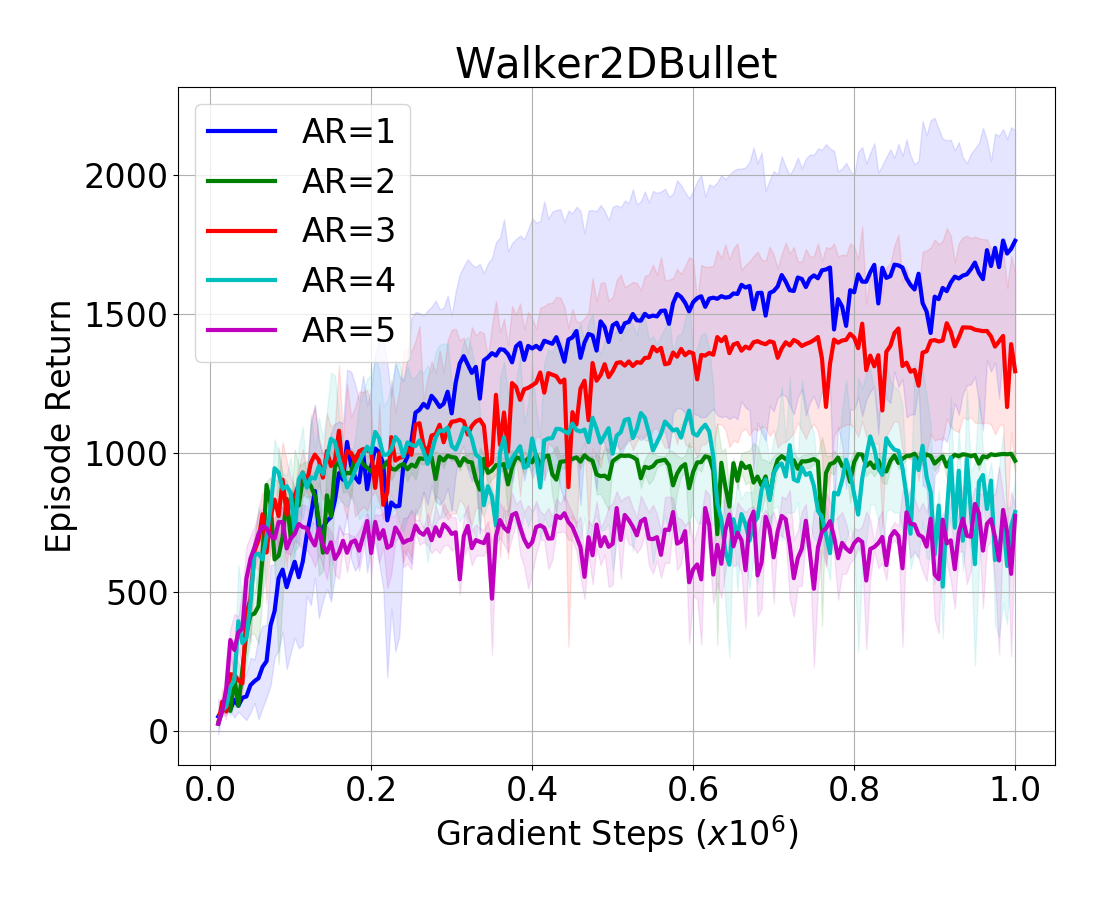}  
  \caption{Walker2DBulletEnv}
  \label{fig:walker_action_repeat}
\end{subfigure}
\begin{subfigure}{\figscale}
  \centering
  \includegraphics[width=\columnwidth]{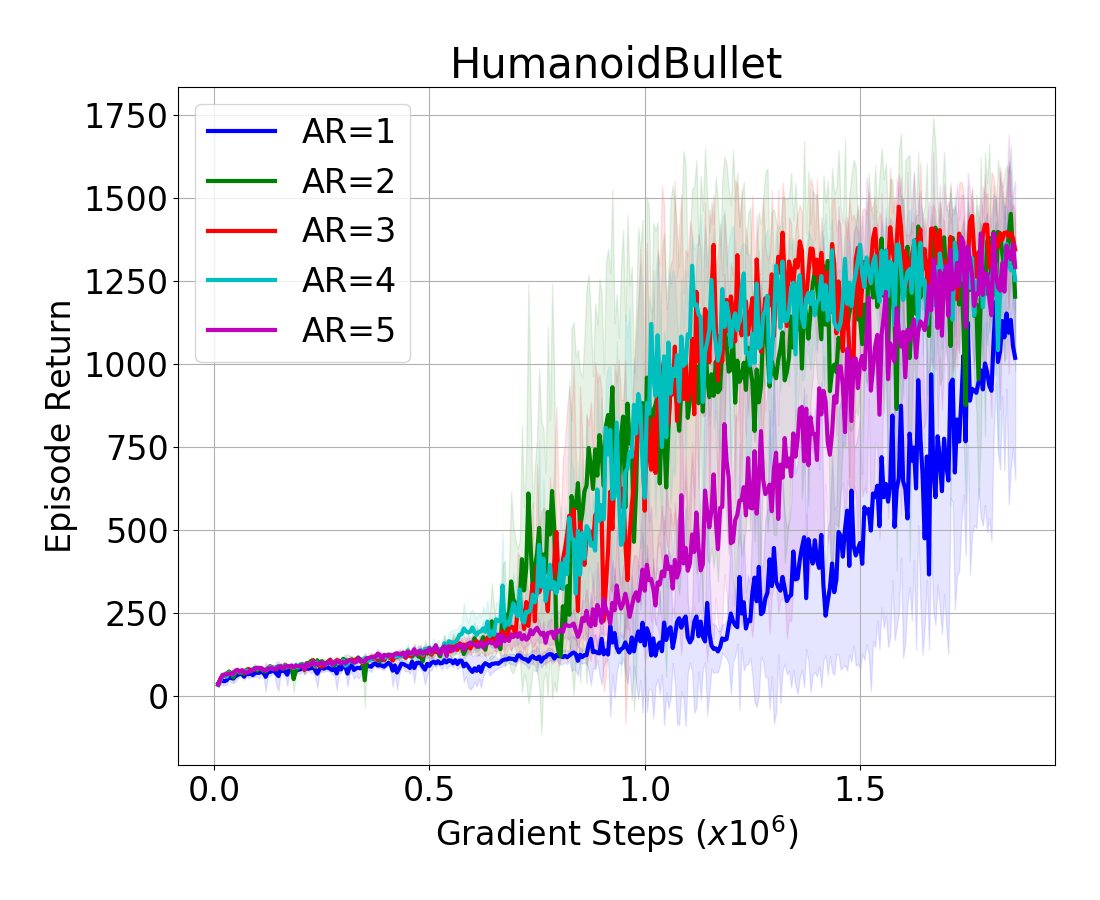}  
  \caption{HumanoidBulletEnv}
  \label{fig:humanoid_action_repeat}
\end{subfigure}
\caption{Impact of Action Repeat.}
\label{fig:action_repeat_results} 
\end{figure}
\section{Episode Termination}

Locomotion environments such as HalfCheetah are constrained with a time limit that 
defines the maximum number of steps per episode. 
The characters continue their exploration until this time limit is reached. 
In both PyBullet~\cite{coumans2019} and Mujoco~\cite{todorov2012mujoco} versions of these environments, the default time limit is 1000. 
In some environments, such as Ant, Hopper, Walker2D, and Humanoid, an additional \textit{natural} 
terminal condition is defined when the character falls. 
In either case, the motivation for limiting the duration of potentially infinite-duration 
trajectories is to allow for diversified experience.

It has been previously shown by~\citet{pardo2018time} that terminations due to time limits should not be 
treated as natural terminal transitions and that these transitions should be bootstrapped instead. 
If we consider a transition at time-step $t$ with starting state $s_t$, action $a_t$, 
transition reward $r_t$, and next state state $s_{t+1}$, with an action-value estimate $\hat{Q}(s_{t+1})$, 
the target value $y$, when bootstrapping, becomes:
\begin{equation*}
y = r_t + \gamma I_{\mathrm{term}} \hat{Q}(s_{t+1}).
\end{equation*}
Here $I_{\mathrm{term}}\in\{0,1\}$ is an indicator variable, set to $0$ for any terminal transition, and is
otherwise $1$, i.e., for non-terminal transitions. 
With infinite bootstrapping, we use $I_{\mathrm{term}}=1$ for time limit terminations,
which corresponds to considering the time limit termination as being non-terminal.

\begin{figure}[bt]
\begin{subfigure}{0.48\columnwidth}
  \centering
 \includegraphics[width=\textwidth]{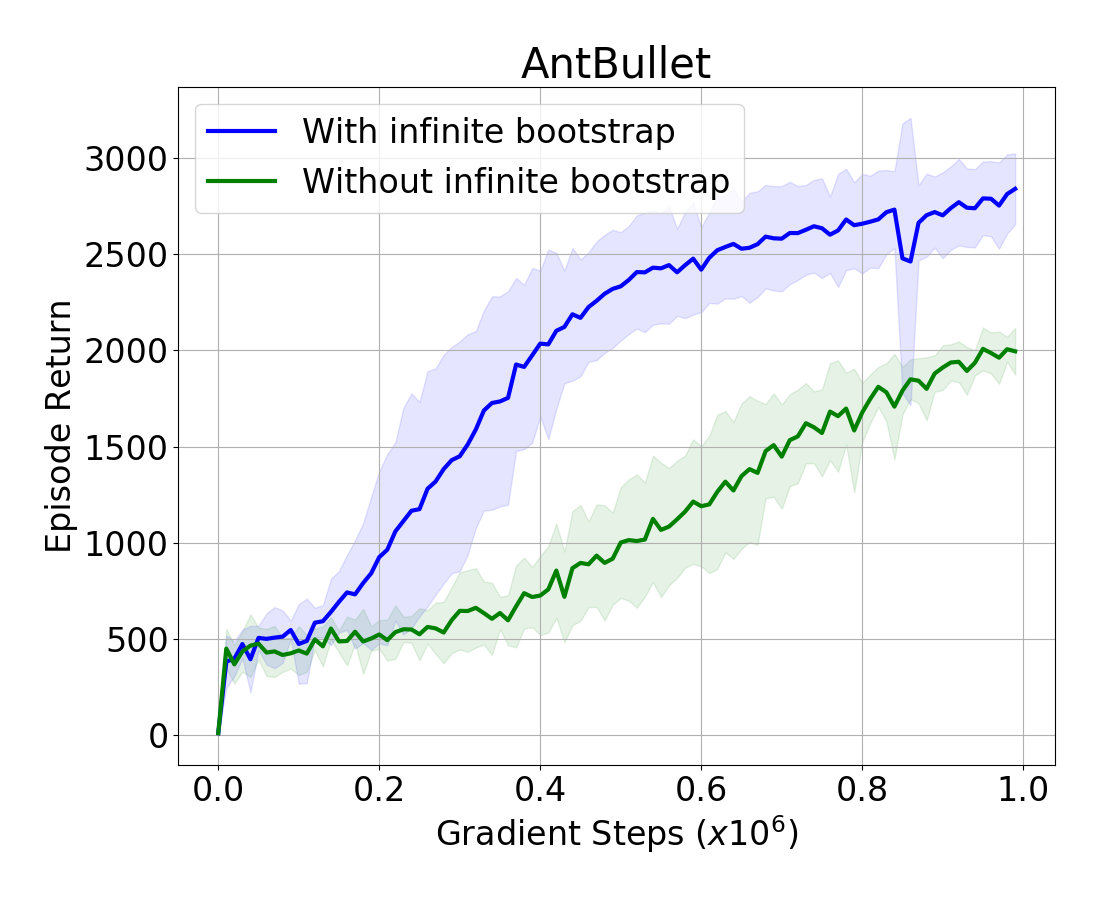}
  \caption{AntBulletEnv}
  \label{fig:ant_infinite_bootstrap}
\end{subfigure}
\begin{subfigure}{0.48\columnwidth}
  \centering
 \includegraphics[width=\textwidth]{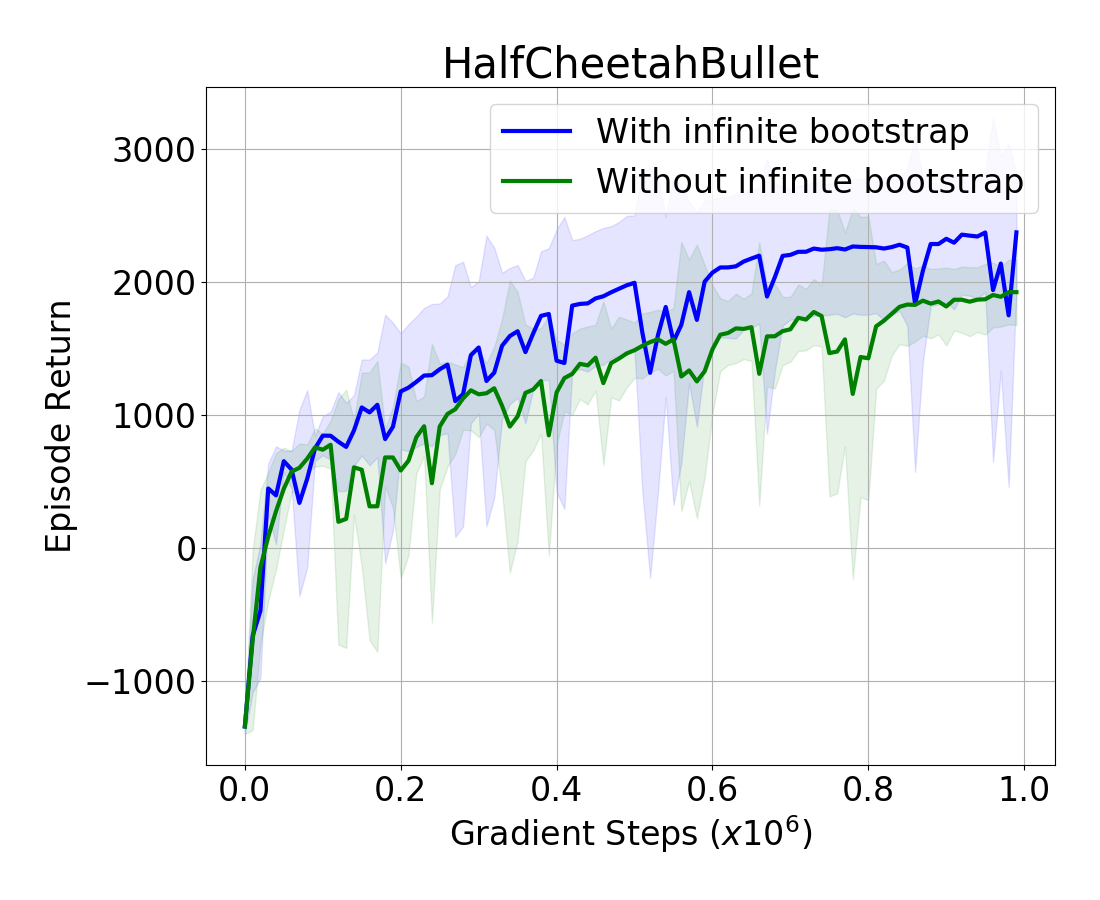}  
  \caption{HalfCheetahBulletEnv}
  \label{fig:halfcheetah_infinite_bootstrap}
\end{subfigure}

\begin{subfigure}{0.48\columnwidth}
  \centering
  \includegraphics[width=\textwidth]{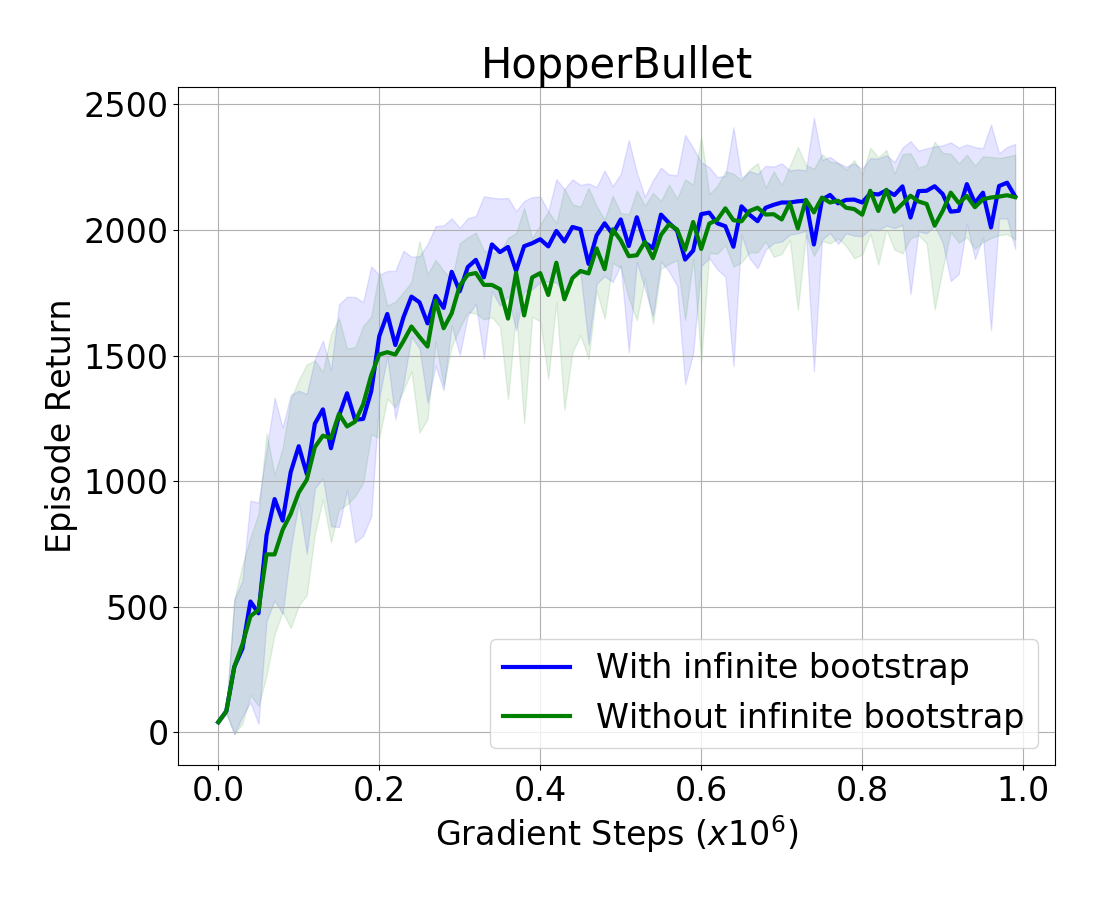}  
  \caption{HopperBulletEnv}
  \label{fig:hopper_infinite_bootstrap}
\end{subfigure}
\begin{subfigure}{0.48\columnwidth}
  \centering
 \includegraphics[width=\textwidth]{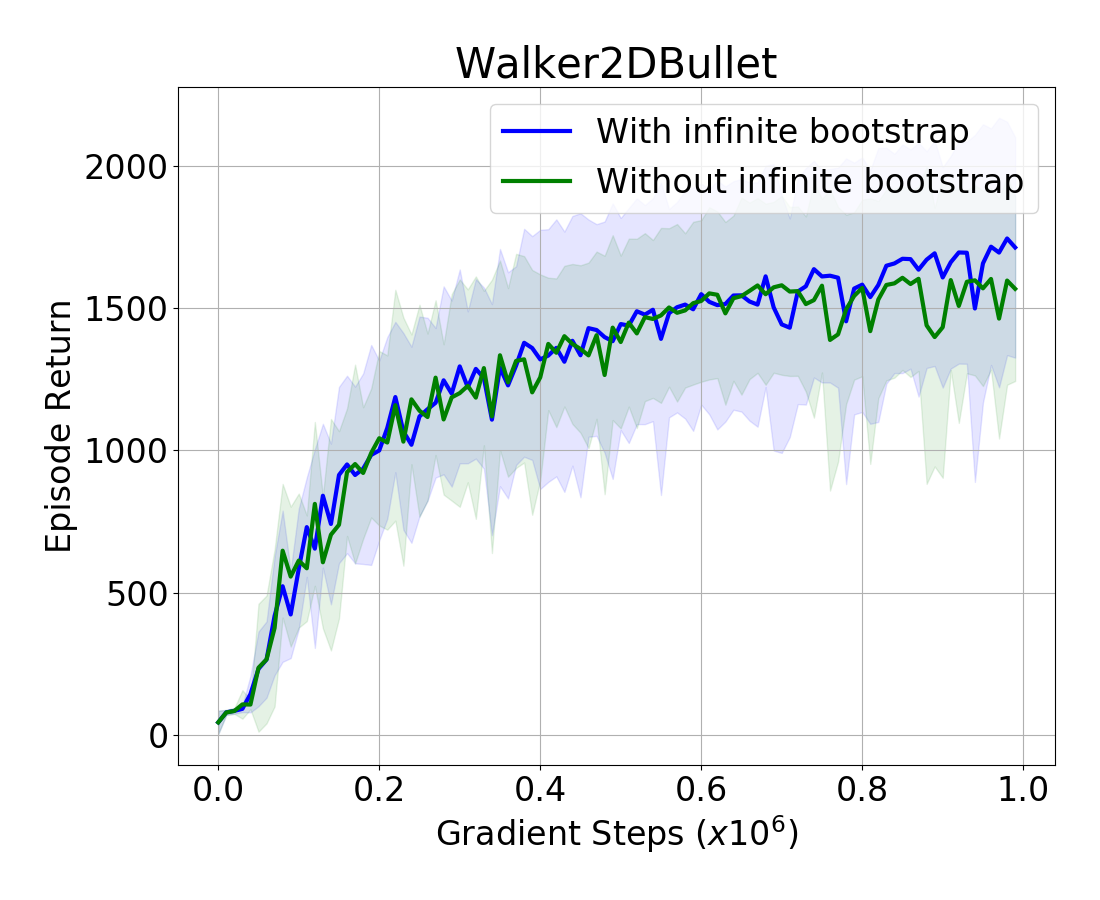}  
  \caption{Walker2DBulletEnv}
  \label{fig:walker_infinite_bootstrap}
\end{subfigure}
\caption{Infinite bootstrap results.}
\label{fig:infinite_bootstrap_results} 
\end{figure}

Our experiments with TD3 are shown in Figure~\ref{fig:infinite_bootstrap_results}. 
These show significant improvements for this infinite bootstrapping trick, in particular for Ant and HalfCheetah,
where the time limit termination is reached more readily.
This finding is consistent with~\citet{pardo2018time} experiments with PPO~\cite{schulman2017proximal}. 
Given that it plays an important role in the final policy and the return value, and that it is often overlooked
or left undocumented, we list it as one of the factors that matters for learning a locomotion policy. 
Like the choice of discount factor, small changes can play an important role in the many environments 
that do not have a natural limited duration.

\textbf{Summary:} Infinite bootstrap is critical for accurately optimizing the action-value function, which usually corresponds to high-quality motion.

\section{Curriculum Learning}
\label{curriculum}

Curriculum-based learning can enable faster learning by progressively
increasing the task difficulty throughout the learning process. These strategies
have been applied to supervised learning~\cite{bengio2009curriculum}
and are becoming increasingly popular in the RL setting, e.g., 
~\cite{florensa2017reverse,yu2018learning,narvekar2019learning,2020-ALLSTEPS}
and many of which are summarized in a recent survey~\cite{narvekar2020curriculum}.
The curriculum therefore serves as a critical source of inductive bias, particularly as the 
task complexity grows. The important problem of curriculum generation is being tackled
from many directions, including the use of RL to learn curriculum-generating policies.
These still face the problem that learning a full curriculum policy can take significantly more experience data 
than learning the target policy from scratch~\cite{narvekar2019learning}.

\textbf{Summary:} By training on task environments with progressively increasing difficulty, 
curriculum learning enables learning locomotion skills that would otherwise be very difficult to learn.



\section{Choice of Action Space}

In most locomotion benchmarks, torque is the dominant actuation model used to drive the articulated character. 
In contrast, low-level stable Proportional-Derivative(PD) controller are used in a variety
of recent results from animation and robotics, e.g.,~\cite{2017-TOG-deepLoco,2019-CORL-cassie,tan2011stable}. 
The PD controller takes in a target joint angle~$\Bar{q}$ as input and outputs torque~$\tau$ according to:
\begin{equation}
    \tau = -k_{p}(q - \Bar{q}) - k_{d}\Dot{q}
    \label{eq:pd_torque}
\end{equation}
where~$q$ and~$\Dot{q}$ represent the current joint angles and velocities, and~$k_{p}$ and~$k_{d}$ are manually defined. 
With a low-level PD controller, the policy produces joint angles rather than torques. 
In previous motion imitation work, e.g., ~\cite{2018-TOG-deepMimic,2019-CORL-cassie,10.1145/3274247.3274506}, the reference trajectory 
provides default values for the PD-target angles, and the policy then learns a residual offset. 
However, this PD-residual policy is infeasible for non-imitation tasks since a reference trajectory is not available. 
Here we study the impact of using (non-residual) PD-action spaces for purely objective-driven 
(vs imitation-driven) locomotion tasks.
For implementation and training details on the PD controller, we direct the reader to appendix~\ref{appendix:pd_control_details}.

Training with low-level PD controllers is more prone to local minimum than the same environment with torque-based RL policy. 
This is because it is relatively easy to learn a constant set of PD-targets which 
maintain a static standing posture, which is rewarded via the survival bonus. 
To address this local-minimum issue, we first train with a broader initial state distribution,
and then, after mastering a basic gait, the agent is trained with the original environment 
with a nearly-deterministic initial-state. 
Among the PyBullet locomotion benchmarks, Walker2D suffers more severely from the problem of a local minimum.

\begin{figure}[tb]
\begin{subfigure}{0.48\columnwidth}
  \centering
  \includegraphics[width=\textwidth]{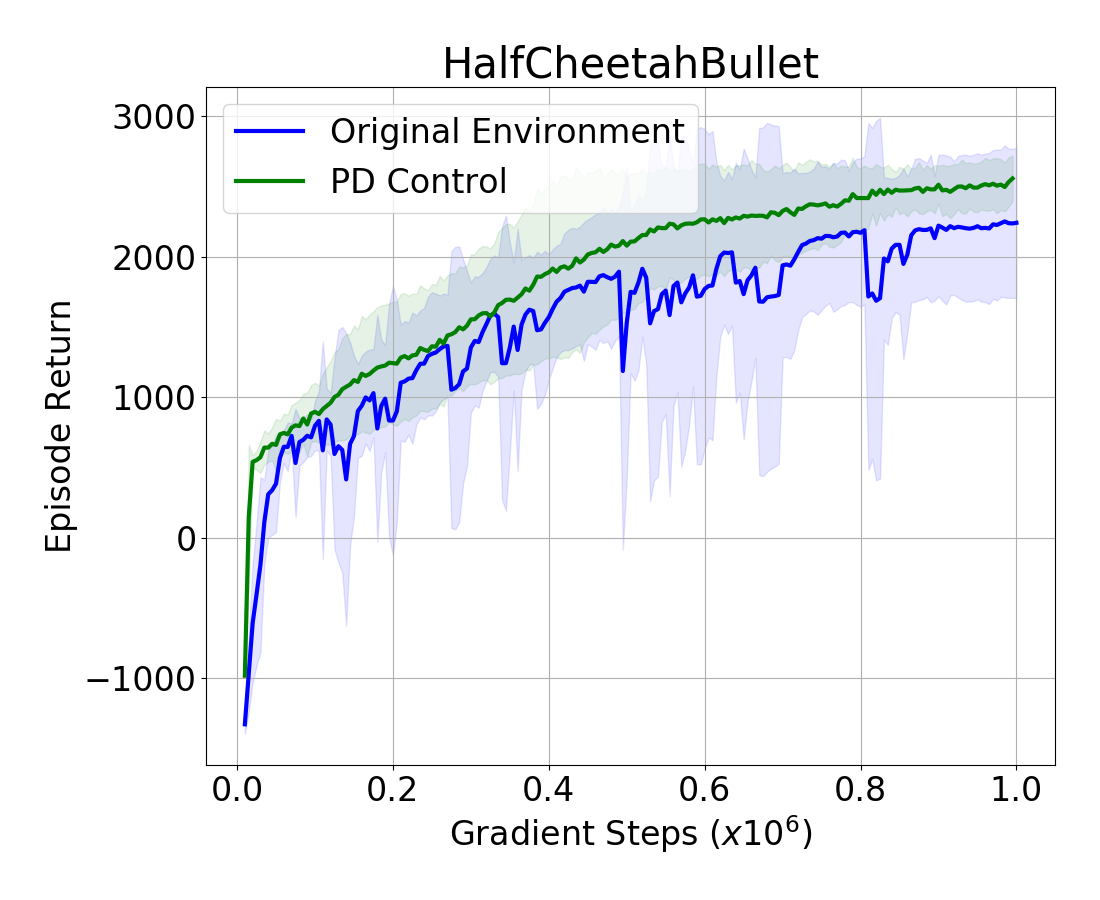}  
  \caption{HalfCheetahBulletEnv}
  \label{fig:halfCheetahBullet_pd}
\end{subfigure}
\begin{subfigure}{0.48\columnwidth}
  \centering
  \includegraphics[width=\textwidth]{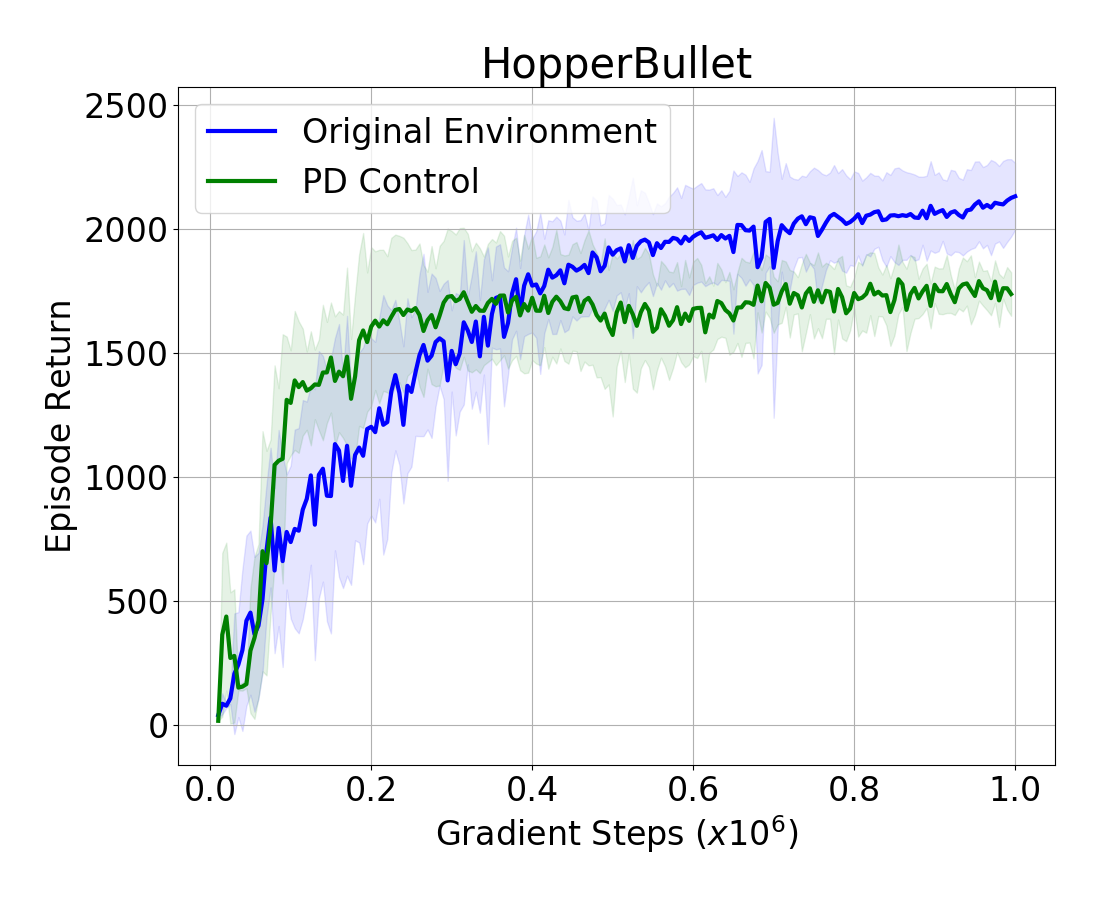}  
  \caption{HopperBulletEnv}
  \label{fig:hopperBullet_pd}
\end{subfigure}
\begin{subfigure}{0.48\columnwidth}
  \centering
  \includegraphics[width=\textwidth]{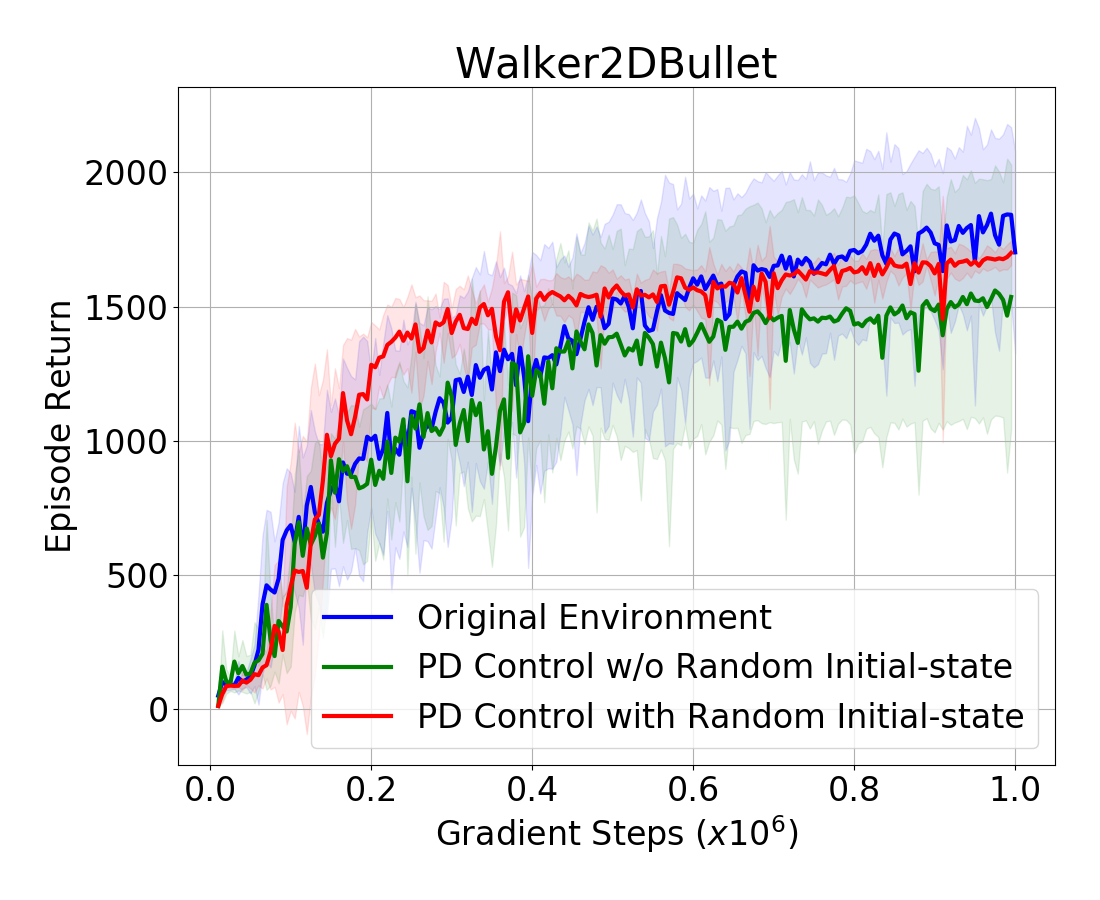}  
  \caption{Walker2DBulletEnv}
  \label{fig:walker2DBullet_pd}
\end{subfigure}
\caption{Comparing learning with PD-control-target and torque action spaces.}
\label{fig:pd_controller_results} 
\end{figure}

As shown in Figure~\ref{fig:pd_controller_results}, using PD target angle as action space often leads 
to a faster early learning rate. 
However, it may converge to lower-reward solutions than the torque action space,
and it may hinder quick adaption to rapid changes in the state 
as the character moves faster~\cite{10.1145/3274247.3274506}.

Previous work~\cite{2017-SCA-action} shows that PD-control-targets outperform torques as a choice of action space
in terms of both final reward and sampling efficiency. 
The PD-control action space is far better at motion imitation tasks than learning locomotion from scratch. 
We believe this arises for two reasons. First. in motion imitation, 
the reference trajectory already serves as a good default, 
and the policy only needs to fine-tune the PD target, which is simpler than learning the PD target from the scratch. 
Second, current locomotion benchmarks are designed and optimized for direct torque control 
rather than low-level PD control, including the state initialization and the reward function.

\textbf{Summary}: For non-imitation locomotion tasks, PD-control-targets actions appears to have 
a limited advantage over a torque-based control strategy. However, 
PD-control-targets often learn a basic gait more efficiently during the early stages of training.
\section{Survival Bonus}
\label{survival_bonus}

The reward function plays an obvious and important role in learning natural and fluid motion. 
In PyBullet locomotion benchmarks, the character is usually rewarded for moving forward with a positive velocity, 
and penalized for control costs, unnecessary collisions, and approaching the joint limit. 
Additionally, the reward function also contains a {\em survival bonus} term, 
which is a positive constant when the character has not fallen; otherwise it is negative or zero. 
In \citet{henderson2018deep} and \citet{mania2018simple}, the existence of the survival bonus terms can 
lead to getting stuck in local minimum for certain types of algorithms. 
Here we investigate the impact of the value of the survival bonus in the reward function.

\begin{figure}[tbh]
  \centering
  \includegraphics[width=\columnwidth]{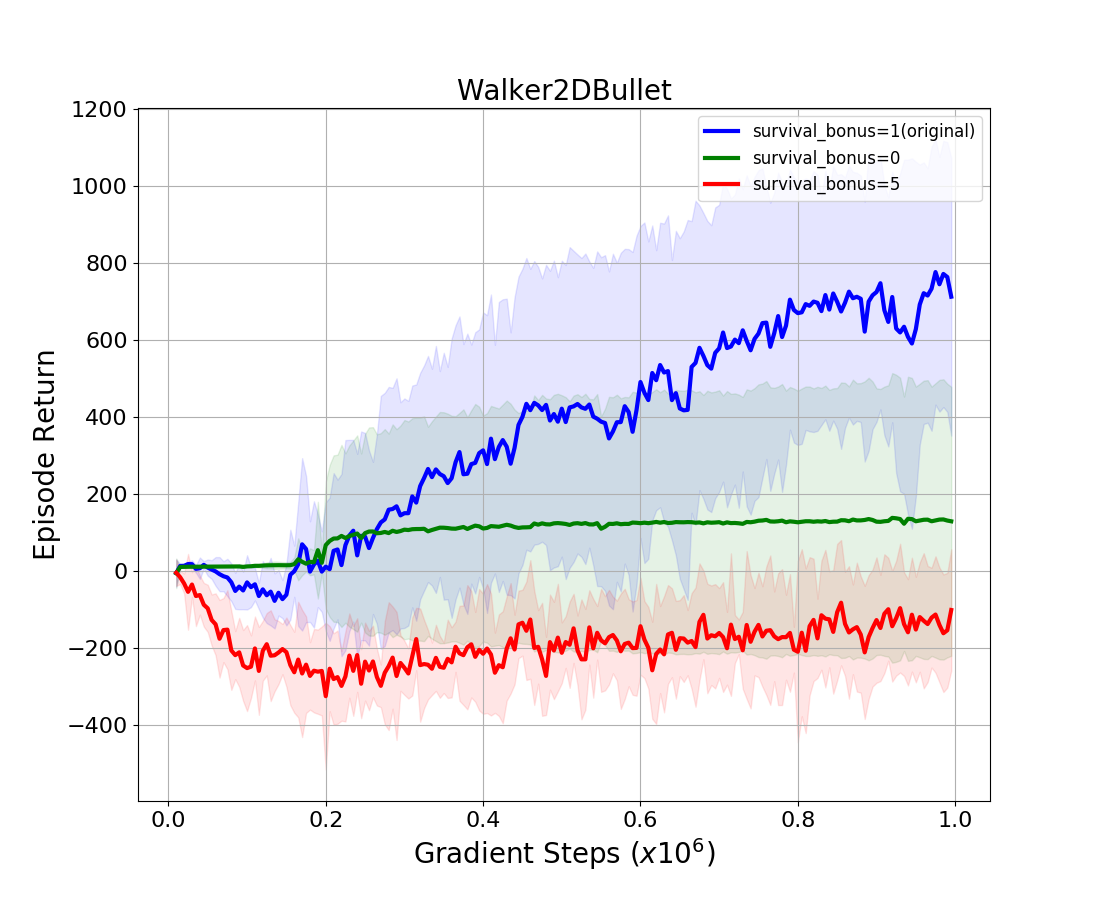}
  \caption{Impact of survival bonus (SB) on Walker2DBulletEnv, evaluated for SB = \{0,1,5\} }
  \label{fig:survival-bonus}
\end{figure}

We experiment with 3 different values of survival bonus: 0, 1 and 5, and train the agent with TD3 algorithm. 
To allow for a fair comparison among policies trained with different survival bonus value, 
the policies are always tested on the environments without the survival bonus reward. 
The default PyBullet environment set the value of survival bonus to 1, 
for which the episodic test reward significantly outperforms the other two tested values for the survival bonus, 
as shown in Figure~\ref{fig:survival-bonus}. 
Including the survival bonus term in the reward function is necessary;
setting it to 0 makes the discovery of a basic walking gait too difficult. 
If the survival bonus term is too large, however, the algorithm exploits the survival bonus reward while neglecting other reward terms.
This results in a character that balances but never steps forward.

\textbf{Summary:} The survival bonus value provides a critical form of reward shaping when learning to locomote.
Values that are too small or too large leads to local minima corresponding to falling-forward and standing still, respectively.

\section{Torque Limit}
\label{torque_limit}

The capabilities of characters are defined in part by their torque limits.
Both high and low torque limits can be harmful to learning locomotion in simulation or actual robots. 
A high torque limit is prone to unnatural behavior. 
On the other hand, a low torque limit is also problematic since it can make it much more difficult to discover
the solution modes that yield efficient and natural locomotion. 
In \citet{Abdolhosseini_2019}, the authors experiment with the impact of torque limits by multiplying the default limits 
by a scalar multiplier $\lambda \in [0.6,1.6]$. Policies are trained using PPO~\cite{schulman2017proximal}. 
The observed result is that higher torque limits obtain higher episodic rewards. 
As $\lambda$ decreases, the agent is more likely to get stuck at local minimum, 
and the agent fails to walk completely when the multiplier is lower than a threshold~\cite{Abdolhosseini_2019}. 
Such results motivate the idea of apply a torque limit curriculum during training,
as a form of {\em continuation method}.  
This allows for large torques early in the learning process, to allow for the discovery
of good solution modes, followed by progressively decreasing torque limits, which then
allows the optimization to find more natural, low torque motions within the good solution modes.

\textbf{Summary:} Large torque limits benefit the exploration needed to find good locomotion modes,
while low torque limits benefit natural behavior. A torque limit curriculum can allow for both.
\section{Conclusions}
\label{discussion}

Reinforcement learning has enormous potential as means of developing physics-based movement
skills for animation, robotics, and beyond. 
Successful application of RL requires not only efficient RL algorithms,
but also the design of suitable RL environments.
In this paper we have investigated a number of issues and design choices related to
RL environments, and we evaluate these on locomotion tasks.
As a caveat, we note that these results have been applied to fairly simple benchmark systems,
and thus more realistic and complex environments may yield different results.
Furthermore, we evaluate these environment design choices using a specific RL algorithm (TD3).
Other RL algorithms may be impacted differently, and we leave that as future work.
However, our work provides a better understanding of the often 
brittle-and-unpredictable nature of RL solutions, as 
bad choices made with regard to defining RL environments quickly become problematic.
Efficient learning of locomotion skills in humans and animals can arguably be attributed in large part to 
their ``RL environment''.  
For example, Underlying reflex based movements and central pattern generators help constrain
the state distribution, provide a suitable action space, and are tuned to control at a particular time scale.

Many environment design issues can be viewed as a form of inductive bias,
e.g., the survival bonus for staying upright, or a given choice of action space.
We anticipate that much of the progress needed for efficient learning of motion skills
will require leveraging the many aspects of RL
that are currently precluded by the canonical environment-and-task structure that is
reflected in common RL benchmarks. As such, state-of-the-art RL-based approaches for animation
should be inspired by common RL algorithms and benchmarks, but should not be constrained by them.

\bibliographystyle{ACM-Reference-Format}
\bibliography{references}

\appendix
\section{Hyperparameters}
\label{appendix:hyperparams}

\begin{table}[ht!]
\centering
\begin{center}
\begin{small}
\begin{tabular}{lc}
\toprule
\bf{Hyperparameter} & \bf{Value} \\
\midrule
Critic Learning Rate & $10^{-3}$ \\
Critic Regularization & None \\
Actor Learning Rate & $10^{-3}$ \\
Actor Regularization & None \\
Optimizer & Adam \\
Target Update Rate ($\tau$) & $5 \cdot 10^{-3}$ \\ 
Batch Size & $100$ \\ 
Iterations per time step & $1$ \\
Discount Factor & $0.99$ \\
Reward Scaling & $1.0$ \\
Normalized Observations & False \\
Gradient Clipping & False \\
Exploration Policy & $\mathcal{N}(0, 0.1)$ \\ 
\bottomrule
\end{tabular}
\caption{Hyperparameters used for training the TD3 algorithm.}
\end{small}
\end{center}
\end{table}

\subsection{TD3 Actor Architecture}

\texttt{nn.Linear(state\_dim, 256)}\\
\texttt{nn.ReLU}\\
\texttt{nn.Linear(256, 256)}\\
\texttt{nn.ReLU}\\
\texttt{nn.Linear(256, action\_dim)}\\
\texttt{nn.tanh}

\subsection{TD3 Critic Architecture}

\texttt{nn.Linear(state\_dim + action\_dim, 256)}\\
\texttt{nn.ReLU}\\
\texttt{nn.Linear(256, 256)}\\
\texttt{nn.ReLU}\\
\texttt{nn.Linear(256, 1)}
\section{Implementation Details of PD Controller}
\label{appendix:pd_control_details}

The PD controller is commonly implemented as an action space wrapper over the original torque based control. To produce stable simulation results, the PyBullet simulator runs at 1200 Hz. Every four substeps, the simulator will receive a torque command to update the dynamics of the character. For each PD target command, the PD controller will convert it to torque iteratively using Equation~\ref{eq:pd_torque} for 5 times such that the low level PD controller runs at 300Hz. The control policy producing PD target angles runs at the same control frequency as the torque policy, 60Hz. The actor outputs the PD target angles as a vector ranged from $[0, 1]$ with the same dimension as the actuator, and then each value from 0 to 1 is mapped to the range bounded by torque limits of each joint.  To obtain a fair comparison with the original environment, the energy cost term and the cost for approaching joint limits term are integrated over the 5 substeps. 

To better avoid local minimum, we set the exploration noise as a Gaussian distribution $\mathcal{N}(0,0.2)$, except for Walker2D environment where the agent is trained on environment with broader initial-state distribution($\kappa$=0.3) for 180K interactions before trained on the default environment.
\section{State definition}
\label{appendix:state_def}

Here we describe the full state definition for the PyBullet environments used in experiments.

A target position $(1000, 0, 0)$ is defined 1000 units away from the root $(0, 0, 0)$.
The default state description in PyBullet environments is composed by the concatenation of the following variables:
\begin{itemize}
    \item change in $z$ coordinate in the world frame, computed as $z_t - z_0$ where $z_t$ is the current position on the $z$ axis and $z_0$ is the initial position (1 real number);
    \item $sin$ and $cos$ of the angle between the character's position and the target position (2 real numbers);
    \item linear velocity of the character with respect to the world frame (3 real numbers);
    \item angle and angular velocity of each joint ($2 \cdot n_{joints}$ real numbers, where $n_{joint}$ is the number of joints, specified in table);
    \item contact information for each foot ($n_{feet}$ binary values, where $n_{feet}$ is described for each environment in the table.
\end{itemize}

\begin{table}[H]
\centering
\begin{center}
\begin{small}
\begin{tabular}{lcc}
\toprule
\bf{Environment Name} & $n_{joint}$ & $n_{feet}$\\
\midrule
HalfCheetahBulletEnv & 6  & 6 \\
AntBulletEnv         & 8  & 4 \\
HopperBulletEnv      & 3  & 1 \\
Walker2DBulletEnv    & 6  & 2 \\
HumanoidBulletEnv    & 17 & 2 \\
\bottomrule
\end{tabular}
\caption{Dimension of the joints and limbs for different characters.}
\end{small}
\end{center}
\end{table}

\end{document}